%% file: arxiv.tex
\pdfoutput=1
\documentclass[10pt,twocolumn]{article}

\usepackage{iccv}
\usepackage{times}
\usepackage{epsfig}
\usepackage{graphicx}
\usepackage{amsmath}
\usepackage{amssymb}
\usepackage[accsupp]{axessibility} 
\input{commands.tex}

\usepackage[pagebackref=true,breaklinks=true,letterpaper=true,colorlinks,bookmarks=false]{hyperref}

\iccvfinalcopy 

\def\iccvPaperID{22} 

\ificcvfinal\fi

\begin{document}

\title{Optical Adversarial Attack}

\author{Abhiram Gnanasambandam, Alex M. Sherman, and Stanley H. Chan\\
Purdue University, West Lafayette, Indiana, USA\\
{\tt\small \{agnanasa,sherma10,stanchan\}@purdue.edu}
}

\maketitle
\ificcvfinal\thispagestyle{empty}\fi

\begin{abstract}
We introduce \textbf{OP}tical \textbf{AD}versarial attack (OPAD). OPAD is an adversarial attack in the physical space aiming to fool image classifiers without physically touching the objects (e.g., moving or painting the objects). The principle of OPAD is to use structured illumination to alter the appearance of the target objects. The system consists of a low-cost projector, a camera, and a computer. The challenge of the problem is the non-linearity of the radiometric response of the projector and the spatially varying spectral response of the scene. Attacks generated in a conventional approach do not work in this setting unless they are calibrated to compensate for such a projector-camera model. The proposed solution incorporates the projector-camera model into the adversarial attack optimization, where a new attack formulation is derived. Experimental results prove the validity of the solution. It is demonstrated that OPAD can optically attack a real 3D object in the presence of background lighting for white-box, black-box, targeted, and untargeted attacks. Theoretical analysis is presented to quantify the fundamental performance limit of the system.
\end{abstract}

\section{Introduction}
\subsection{What is OPAD?}
Adversarial attacks and defenses today are predominantly driven by studies in the \emph{digital} space \cite{szegedy_13_intro_attack, goodfellow_14_explaining_adversarial, li_14_feature, moosavi_17_universal, nguyen_15_deep_fooled, papernot_16_transferability, sabour_15_adversarial,bhattad_19_color_att, gu2014towards, papernot_16_transferability,feinman2017detecting,grosse2017statistical,lu2017safetynet,metzen2017detecting} where the attacker manipulates a digital image on a computer. The other form of attacks, which are the \emph{physical} attacks, have been reported in the literature \cite{morgulis_19_traffic_replace,tu_2020_lidar_attack,cao_19_lidar,kurakin_16_phy_adversarial,eykholt_18_stopsign_phy, song_18_physical_STOP,worzyk_19_uncompensated_projector,xu2019adversarial}, but most of the existing ones are invasive in the sense that they need to touch the objects, for example, painting a stop sign \cite{eykholt_18_stopsign_phy}, wearing a colored shirt \cite{xu2019adversarial}, or 3D-printing a turtle \cite{athalye_18_synthesizing}. In this paper, we present a non-invasive attack using structured illumination. The new attack, called the \textbf{OP}tical \textbf{AD}versarial attack (OPAD), is based on a low-cost projector-camera system where we project calculated patterns to alter the appearance of the 3D objects.

The difficulty of launching an optical attack is making sure that the perturbations are imperceptible while compensating for the environmental attenuations and the instrument's nonlinearity. An optimal attack pattern in the digital space can become a completely different pattern when illuminated in the real 3D space because of the background lighting, object reflectance, and nonlinear response of the light sources. OPAD overcomes these difficulties by taking into consideration the environment and the algorithm. OPAD is a meta-attack framework that can be applied to any existing digital attack. The uniqueness and novelty of OPAD are summarized in three aspects:
\begin{itemize}
\setlength\itemsep{0ex}
\item OPAD is the first method in the literature that explicitly models the instrumentation and the environment. Thus, the adversarial loss function in the OPAD optimization is interpretable and is transparent to the users.
\item Most of the illumination-based attacks in the literature require iteratively capturing and optimizing for the attack patterns. OPAD is non-iterative. It attacks real 3D objects in a single shot. Furthermore, it can launch targeted, untargeted, white-box, and black-box attacks.
\item OPAD has a theoretical guarantee. With OPAD, we know exactly what objects can be attacked and what cannot. We know the smallest perturbation that is required to compensate for the environmental and instrumental attenuations.
\end{itemize}

\begin{figure*}[!]
\centering
\begin{tabular}{c}
\includegraphics[width = 0.9\linewidth]{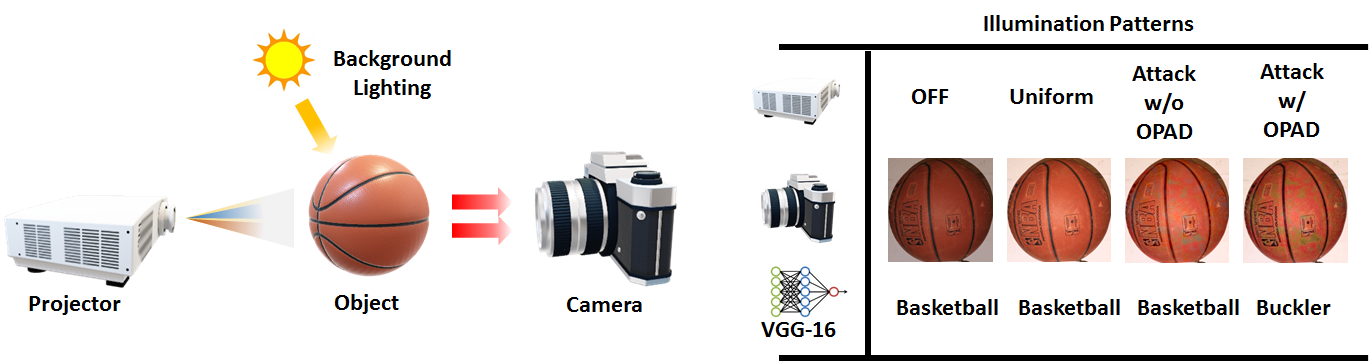}
\end{tabular}
\caption{OPAD is a projector-camera system that uses structured illumination to perturb the appearance of objects. The four configurations shown on the right-hand side are: (1) When the projector is off, the classification remains correct. (2) Whenever we turn on the projector, it will generate a uniform illumination. Most classifiers today are robust to this kind of brightness offset. (3) If we illuminate the object with an attack pattern but do not compensate for the environment loss, the classification remains correct. (4) When we use OPAD to compensate for the loss, we successfully attack the classifier to the targeted class.}
\label{fig:prob_desc}
\end{figure*}

\begin{figure*}[ht]
\centering
\begin{tabular}{ccc}
\hspace{-2.0ex} Uniform illumination & \hspace{-2.0ex} Our illumination & \hspace{-2.0ex} Captured \\
\includegraphics[height = 3.7cm]{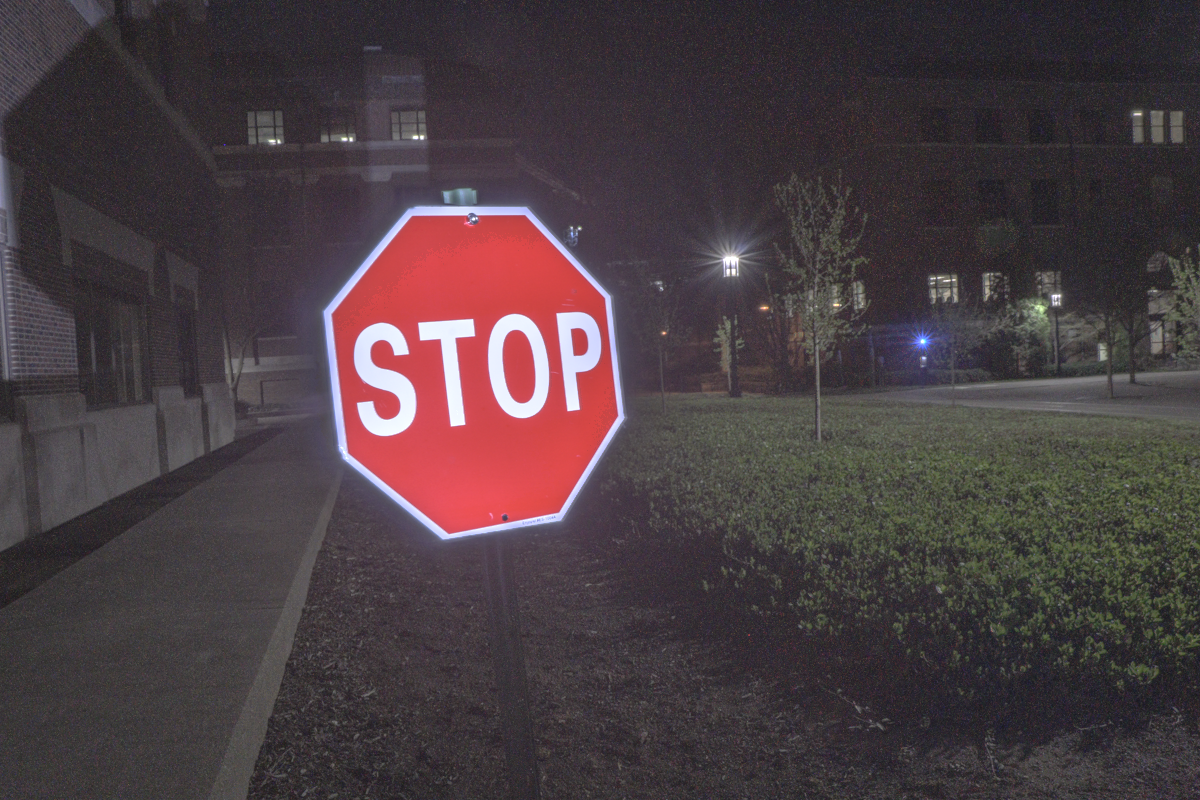}&
\hspace{-2.0ex}\includegraphics[height = 3.7cm]{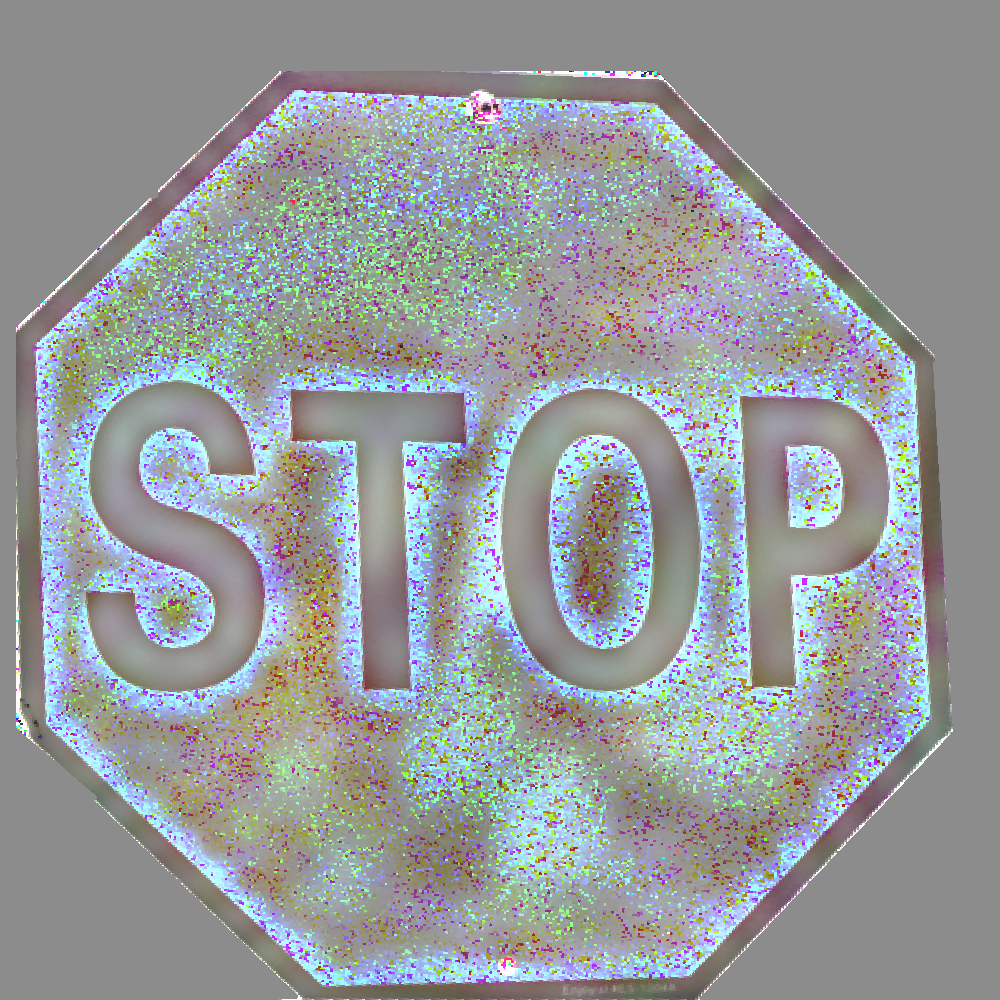}&
\hspace{-2.0ex}\includegraphics[height = 3.7cm]{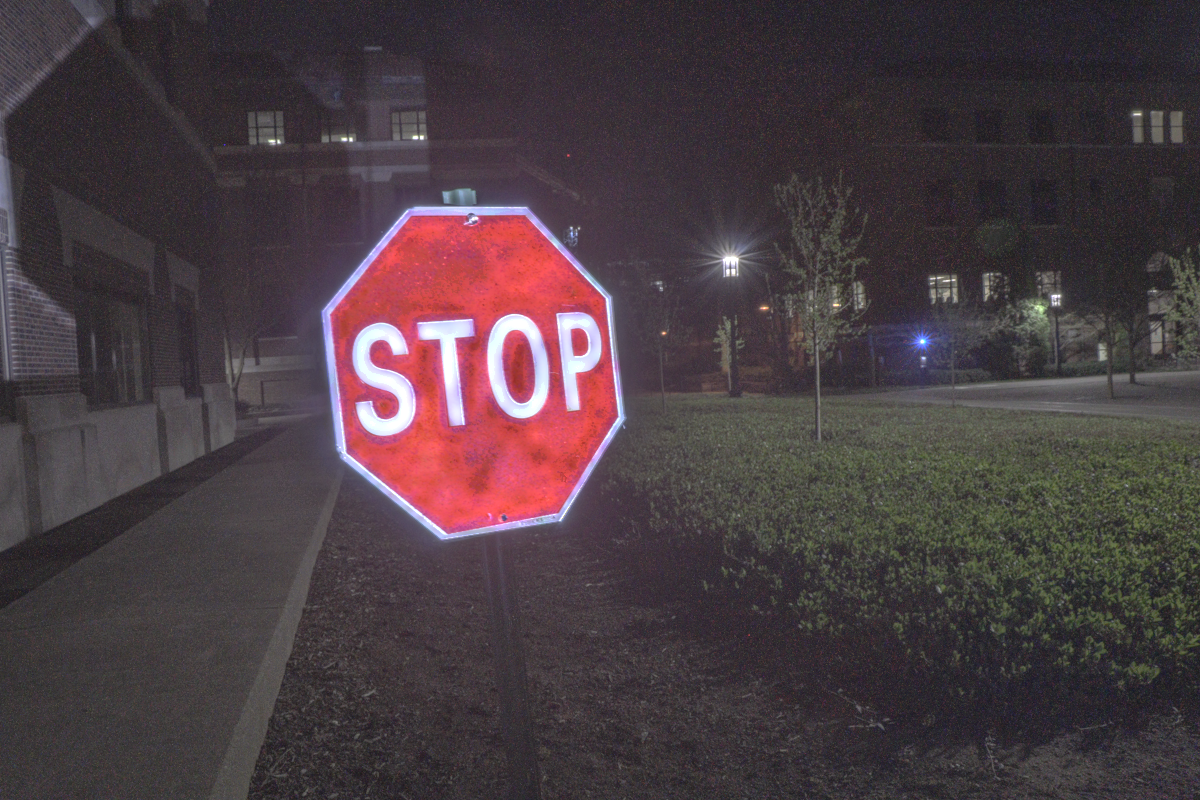}\\
Stop Sign &\hspace{-2.0ex} &\hspace{-2.0ex} Speed 30
\end{tabular}
\caption{An actual optical setup for OPAD. In this experiment, we attack a real STOP sign. The baseline image is obtained by illuminating the object with a uniform illumination of an intensity 140/255. To attack the object, we generate a projector-compensated illumination with Madry et al. \cite{madry2017towards} ($\ell_\infty$ projected gradient descent attack) as the backbone. When projecting this structured illumination onto the metallic stop sign, the prediction becomes Speed 30.}
\label{fig:STOPsign_white_box}
\end{figure*}

To provide a preview of the proposed OPAD framework, in \fref{fig:prob_desc} we show a schematic diagram and four scenarios. The OPAD system consists of a projector and a camera. When the projector is off, the camera sees the unperturbed object. This is the vanilla baseline used by a conventional digital attack. When the projector is turned on, it will project a uniform pattern onto the object. This is the new baseline. Note that this new baseline is required for all-optical perturbations. As long as an active light source exists, a constant offset will be introduced through the uniform illumination. Most classifiers are not affected by such an offset. The interesting phenomenon happens when we project a digital attack pattern (e.g., FGSM \cite{goodfellow_14_explaining_adversarial}). Since we have not compensated for the projector's nonlinearity, the attack will be attenuated. Thus, the object will still be classified correctly. With OPAD, we compensate for the environmental and instrumental distortions. The new perturbation can mislead the classifier. \fref{fig:STOPsign_white_box} shows our proof-of-concept OPAD system in a real outdoor scene. The system consists of a ViewSonic 3600 Lumens SVGA projector, a Canon T6i camera, and a laptop computer. We showed how OPAD could make the metallic stop sign be classified as Speed 30.

\subsection{Related work}
The scope of the paper belongs to optics-based attacks. The reported results in the literature are few \cite{worzyk_19_uncompensated_projector,AAAI_attack,CVPRW_attack,duan2021adversarial} and there are many limitations.

\noindent$\bullet$ \textbf{Iterative approaches} \cite{worzyk_19_uncompensated_projector,AAAI_attack}: These methods do not consider the forward model of the instrument and environment. Thus they need to capture images and calculate the attack iteratively. OPAD is a single-shot attack.

\noindent$\bullet$ \textbf{Attack a displayed image} \cite{worzyk_19_uncompensated_projector}: Attacks of this type cannot attack real 3D objects. OPAD can attack 3D objects.

\noindent$\bullet$ \textbf{Un-targeted attack with unbounded magnitude} \cite{duan2021adversarial}: One can use a strong laser to create a beam in the scene. However, such an attack is un-targeted, and the magnitude of the perturbation is practically unbounded. Perturbations of OPAD are targeted, and they are minimally perceptible.

\noindent$\bullet$ \textbf{Global color correction methods} \cite{CVPRW_attack}: Methods based on this principle have limited generality because the optics are spatially varying and spectrally nonlinear. OPAD explicitly takes into account these considerations.

A key component OPAD builds upon is the prior work of Grossberg et al. \cite{grossberg_04_ShreeNayar_calibration} (and follow-ups \cite{huang_20_new_calibration, siegl_17_adaptive}). Although in a different context, Grossberg et al. demonstrated a principled way to compensate for the loss induced by the projector. OPAD integrates the model with the adversarial attack loss maximization. This new combination of optics and algorithms is new in the literature.

\section{Projector-camera model}
OPAD is an integration of a projector-camera model and a loss maximization algorithm. In this section, we discuss the projector-camera model.

\subsection{Notation}
We use $x \in \R^2$ to denote the 2D coordinate of a digital image. The $x$-th pixel of the source illumination pattern being sent to the projector is denoted as $\vf (x) = [f_R(x),\; f_G(x),\; f_B(x)]^T \in \R^3$, and the overall source pattern is $\mathbf{f} = [\vf(x_1),\vf(x_2),\ldots,\vf(x_N)]^T \in \R^{3N}$, where $N$ is the number of pixels.

As the source pattern goes through the projector and is reflected by the scene, the actual image captured by the camera is
\begin{equation}
    \mathbf{g} = \calT(\mathbf{f}),
\end{equation}
where $\mathbf{g} \in \R^{3N}$ is the observed image, and $\calT: \R^{3N} \rightarrow \R^{3N}$ is the overall mapping of the forward model. To specify the mapping at pixel $x$, we denote $\calT^{(x)}: \R^3 \rightarrow \R^3$ with $\vg(x) = \calT^{(x)}(\vf(x))$, or simply $\vg(x) = \calT(\vf(x))$ if the context is clear.

\subsection{Radiometric response}
\begin{figure}[ht]
\centering
\includegraphics[width=\linewidth]{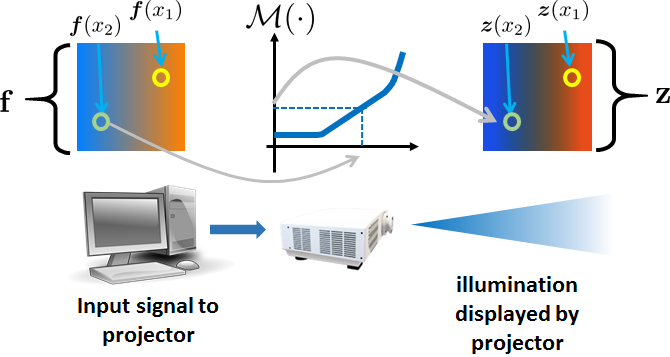}
\caption{Radiometric response of a projector. As we send an input illumination from the computer to the projector, the input signal $\mathbf{f}$ is altered by the radiometric response function $\calM(\cdot)$. $\calM(\cdot)$ is performed per color channel per pixel. The actual signal displayed by the projector is $\mathbf{z}$.}
\label{fig: opad projector radiometric}
\end{figure}

As the source pixel $\vf(x) \in \R^3$ is sent to the projector, the nonlinearity of the projector will alter the intensity per color channel. This is done by a radiometric response function $\calM = [\calM_R, \calM_G, \calM_B]^T$ which transforms the desired signal $\vf(x)$ to a projector brightness signal $\vz(x) \in R^3$:
\begin{equation*}
\vz(x) \bydef
\begin{bmatrix}
z_R(x)\\
z_G(x)\\
z_B(x)
\end{bmatrix}
=
\begin{bmatrix}
\calM_R(f_R(x))\\
\calM_G(f_G(x))\\
\calM_B(f_B(x))
\end{bmatrix}
= \calM(\vf(x)).
\end{equation*}
This is illustrated in \fref{fig: opad projector radiometric}.

Every projector has its own radiometric response. This is intrinsic to the projector, but it is independent of the scene.

\subsection{Spectral response}
\begin{figure}[ht]
\centering
\includegraphics[width=\linewidth]{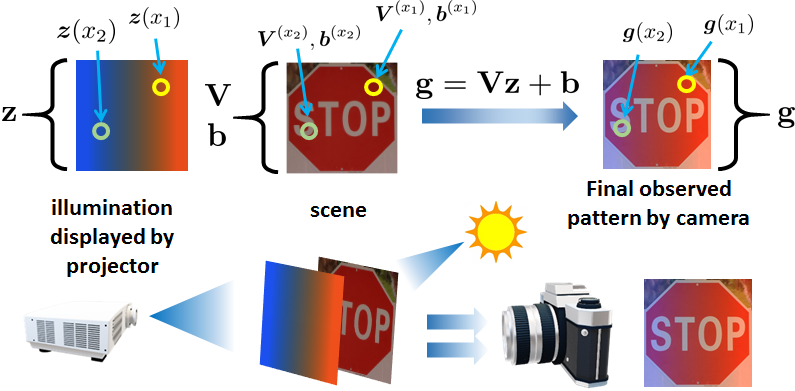}
\caption{Spectral response of a projector. Given the illumination pattern displayed by the projector, the scene and background lighting will be applied to the illumination via a color transform matrix $\mV^{(x)}$ and an offset vector $\vb^{(x)}$. The final image captured by the camera follows \eref{eq: g = Vx+b}.}
\label{fig: opad projector spectral}
\end{figure}

The second component of the camera-projector model is the spectral response due to the irradiance and reflectance of the scene. This converts $\vz(x)$ to the observed pixel $\vg(x)$ using a color transform. Since the spectral response encodes the scene, and the color transform is spatially varying,
\begin{equation}
    \vg(x) = \mV^{(x)}\vz(x) + \vb^{(x)},
    \label{eq: g = Vx+b}
\end{equation}
where $\mV^{(x)}$ is a $3 \times 3$ color mixing matrix defined as
\begin{equation}
    \mV^{(x)}=
    \begin{bmatrix}
V_{RR}^{(x)} & V_{RG}^{(x)} & V_{RB}^{(x)}\\
V_{GR}^{(x)} & V_{GG}^{(x)} & V_{GB}^{(x)}\\
V_{BR}^{(x)} & V_{BG}^{(x)} & V_{BB}^{(x)}
\end{bmatrix}.
\label{eqn:colormixing}
\end{equation}
Here, the superscript $(\cdot)^{(x)}$ emphasizes the spatially varying nature of the matrix $\mV^{(x)}$, whereas the subscript clarifies the color mixing process from one input color to another output color. The vector $\vb^{(x)}$ is an offset accounting for background illumination. It is defined as $\vb^{(x)} = [b_R^{(x)}; b_G^{(x)}; b_B^{(x)} ] \in \R^3$. A schematic diagram illustrating the spectral response is shown in \fref{fig: opad projector spectral}.

If the input illumination is $\mathbf{f}$, the final output observed by the camera is
\begin{equation}
    \mathbf{g} =
    \underset{\calT(\mathbf{f})}{\underbrace{\mathbf{V}\calM(\mathbf{f})+\mathbf{b}}},
    \label{eq: T}
\end{equation}
where $\mathbf{V} = \text{diag}\{\mV^{(x_1)},\mV^{(x_2)},\ldots,\mV^{(x_{N})}\} \in \R^{3N \times 3N}$ is a block diagonal matrix where each block $\mV^{(x_{n})}$ is a 3-by-3 matrix. The mapping $\calM$ is an elementwise transform representing the radiometric response. The vector $\mathbf{b} = [\vb^{(x_1)},\ldots,\vb^{(x_N)}]^T \in \R^{3N}$ is the overall offset.

The estimation of both the radiometric and the spectral response is discussed in the supplementary material.

\section{OPAD Algorithm}
\begin{figure}[ht]
\centering
\includegraphics[width=\linewidth]{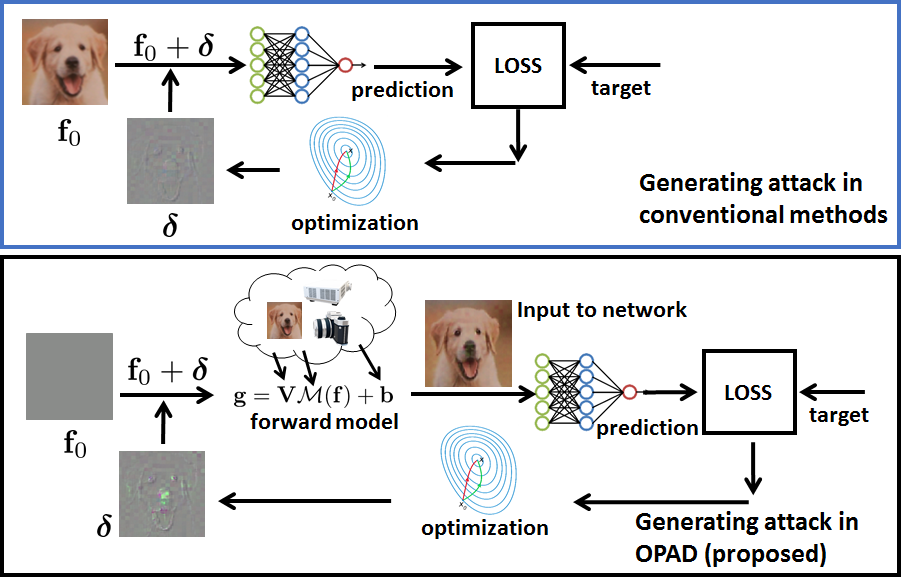}
\caption{In conventional digital attack, the perturbation is directly added to the input image by computing the gradient of the network. In OPAD, the base input $\mathbf{f}_0$ is the uniform illumination. Perturbation $\vdelta$ is added to $\mathbf{f}_0$. The projector and the scene kick in through the radiometric response and the spectral response, respectively.}
\label{fig: OPAD optimization}
\end{figure}

OPAD is a meta procedure that can be applied to any existing adversarial loss maximization. Because the radiometric and spectral response of the projector-camera system treats the illumination and the scene in two different ways, the loss maximization in OPAD is also different from a conventional attack as illustrated in \fref{fig: OPAD optimization}.

\subsection{OPAD loss maximization}
For simplicity we formulate the targeted white-box attack. Other forms of attacks (black-box, and/or untargeted) can be derived similarly. Consider a uniform illumination pattern $\mathbf{f}_0$ that gives a clean image $\mathbf{g}_0 = \mathbf{V}\calM(\mathbf{f}_0)+\mathbf{b}$. Our goal is to make the classifier think that the label is $\ell_{\text{target}}$. The white-box attack is given by
\begin{align}
    \vdelta
    &= \argmax{\vdelta}\;\;  \calL(\calT(\mathbf{f}_0 + \vdelta), \ell_{\text{target}}) \notag \\
    &= \argmax{\vdelta} \;\; \calL( \mathbf{V}\calM(\mathbf{f}_0 + \vdelta) + \mathbf{b}, \ell_{\text{target}}).
\end{align}

In most of the attack methods, the attack $\vdelta$ is constrained in the \emph{input} space through an $\epsilon$-ball such as $\|\vdelta\|<\epsilon$. This only ensures that the input is similar before and after the attack. In our problem, we are interested in two constraints:
\begin{itemize}
    \item The attack in the \emph{output} space should have a small magnitude so that the displayed images before and after attack are visually similar. That is, we want
    \begin{equation}
        \bigg\|
        \underset{{\bydef\mathbf{g}_0}}{\underbrace{\left(\mathbf{V}\calM(\mathbf{f}_0) + \mathbf{b}\right)}} -
        \underset{{\bydef\mathbf{g}}}{\underbrace{\left(\mathbf{V}\calM(\mathbf{f}_0 + \vdelta) + \mathbf{b}\right)}} \bigg\| < \alpha,
    \end{equation}
    for some upper bound constant $\alpha$.
    \item The perturbed illumination has to be physically achievable, meaning that
    \begin{equation}
        0 \le \mathbf{f}_0 + \vdelta \le 1.
    \end{equation}
\end{itemize}

Putting these constraints into the formulation, the attack is obtained by solving the optimization
\begin{align}
    \vdelta^*
    = &\argmax{\vdelta}\;\; \calL( \mathbf{V}\calM(\mathbf{f}_0 + \vdelta) + \mathbf{b}, \ell_{\text{target}}) \notag\\
    &\mbox{subject to} \;\; \notag\\
    &   \|\left(\mathbf{V}\calM(\mathbf{f}_0) + \mathbf{b}\right) - \left(\mathbf{V}\calM(\mathbf{f}_0 + \vdelta) + \mathbf{b}\right)\| < \alpha. \notag\\
    &   0 \le \mathbf{f}_0 + \vdelta \le 1. \tag{P1} \label{eqn:P1}
\end{align}

\subsection{Simplifying the formulation}
Solving \eref{eqn:P1} is challenging because it involves computing $\mathbf{V}$ and $\calM$ in a nonlinear way. However, it is possible to simplify the problem. Using $\mathbf{g}$ and $\mathbf{g}_0$ defined in the first constraint, we can define a perturbation $\veta$ in the \emph{output} space as
\begin{equation}
    \veta \bydef \mathbf{g}-\mathbf{g}_0.
\end{equation}
Substituting this into \eref{eqn:P1}, we can rewrite the problem as
\begin{align}
    \veta^*
    = &\argmax{\veta}\;\; \calL( \mathbf{g}_0+\veta, \ell_{\text{target}}) \notag\\
    &\mbox{subject to} \qquad \| \veta\| < \alpha, \notag\\
    &\qquad\qquad   0 \le \mathbf{f}_0 + \vdelta \le 1.
\end{align}
Thus, it remains to rewrite the second constraint. To this end, we notice that if we apply $\calM$ and $\mathbf{V}$ to $\mathbf{f}_0+\vdelta$, we will alter the box constraint $0 \le \mathbf{f}_0 + \vdelta \le 1$ (which is equivalent to $\vc_\ell \le \vdelta \le \vc_{u}$ where $\vc_\ell = -\mathbf{f}_0$ and $\vc_u = 1-\mathbf{f}_0$) to a new constraint set:
\begin{align*}
    \Omega =
    \bigg\{\veta
    \;\bigg|\; &\veta = \mathbf{V}\calM(\mathbf{f}_0 + \vdelta) - \mathbf{V}\calM(\mathbf{f}_0), \;\; \vc_\ell \le \vdelta \le \vc_u \bigg\}.
\end{align*}
As we will derive below, this constraint is met by projecting the current estimate onto $\Omega$. Putting everything together, we arrive at the final attack formulation:
\begin{align}
    \veta^*
    = &\argmax{\veta \in \Omega}\;\; \calL( \mathbf{g}_0+\veta, \ell_{\text{target}}) \notag\\
    &\mbox{subject to} \qquad \| \veta\| < \alpha. \tag{P2} \label{eqn:P2}
\end{align}

\subsection{OPAD procedure}
If we ignore the constraint set $\Omega$ for a moment, \eref{eqn:P2} is a standard attack optimization that can be solved in various ways, e.g., fast gradient sign method (FGSM) \cite{goodfellow_14_explaining_adversarial}, projected gradient descent (PGD) \cite{madry2017towards}, and many others \cite{bhattad_19_color_att,carlini_17_robustness,kurakin_16_phy_adversarial,he2016deep}. The per-iteration update of these algorithms can be written in a generic form as
\begin{equation}
    \veta^{t+1} = \text{my attack}(\mathbf{f_0}, \veta^{t}, \ell),
\end{equation}
where `$\text{my attack}(\cdot)$' can be chosen as any of the attacks listed above. For example if we use PGD with $\ell_\infty$ constraint, then $\veta^{t+1} =\alpha \cdot \text{sign} \{ \nabla \calL( \vg_0 + \veta^t , \ell_\text{target}) \}$.

In the presence of the constraint set $\Omega$, the per-iteration update will involve a projection:
\begin{equation}
    \veta^{t+1} = \text{Project}_{\Omega} \bigg\{
    \underset{{=\veta^{t+\frac{1}{2}}}}{\underbrace{\text{my attack}(\mathbf{f_0}, \veta^{t}, \ell_\text{target})}}\bigg\}.
\end{equation}
Specific to our problem, the projection operation inverts the current estimate from the output space to the input space and do the clipping in the input space. Then, we re-map the signal back to the output space. Mathematically, the projection is defined as
\begin{equation}
\text{Project}_{\Omega}(\veta^{t+\frac{1}{2}}) = \calT\left( \left[\calT^{-1}\left(\mathbf{g}_0 + \veta^{t+\frac{1}{2}}\right)\right]_{[0,1]} \right) - \mathbf{g_0},
\end{equation}
where $\calT$ is the forward mapping defined in \eref{eq: T} and $[\;\cdot\;]_{[0,1]}$ means clipping the signal to $[0,1]$.

For implementation, the overall attack is estimated by first running an off-the-shelf adversarial attack for one iteration. Then we use the pre-computed projector-camera model $\calT$ (and its inverse $\calT^{-1}$) to handle the constraint set $\Omega$. Since $\mathbf{V}$ is just a matrix vector multiplication, and $\calM$ is a pixel-wise mapping (which can be stored as a look-up table), the overall computational cost is comparable to the original attack.

\section{Understanding the geometry of OPAD}
We analyze the fundamental limit of OPAD by considering linear classifiers. Consider a binary classification problem with a true label $\ell_{\text{true}} \in \{+1,-1\}$. We assume that the classifier $h: \R^{3N} \rightarrow \{+1,-1\}$ is linear, so that the prediction is given by
\begin{equation}
    \widehat{\ell}_{\text{predict}} = h(\mathbf{g}_0) = \text{sign}{(\vtheta^T \mathbf{g}_0)},
\end{equation}
where $\vtheta$ is the classifier's parameter, and $\mathbf{g}_0 = \mathbf{V}\calM(\mathbf{f}_0) + \mathbf{b}$ is the clean image generated by the lower pipeline of \fref{fig: OPAD optimization}. The loss function $\calL_{\vtheta}(\cdot)$ for this sample $\mathbf{g}_0$ is
\begin{equation}
    \calL_{\vtheta}(\mathbf{g}_0,\ell_{\text{true}}) = -\ell_{\text{true}} \cdot \vtheta^T \mathbf{g}_0
\end{equation}

Suppose that we attack the classifier by defining $\mathbf{g} = \mathbf{g}_0 + \veta$. Then, the loss function becomes
\begin{equation}
    \calL_{\vtheta}(\mathbf{g},\ell_{\text{target}}) = -\ell_{\text{target}} \cdot \vtheta^T \left(\mathbf{g}_0 + \veta\right).
    \label{eq: loss function}
\end{equation}
Substituting \eref{eq: loss function} into \eref{eqn:P2}, we show that
\begin{align}
    \veta^*
    &= \argmax{\veta} \;\; -\ell_{\text{target}} \cdot \vtheta^T\veta, \notag\\
    &\text{subject to} \;\; \veta \in \Omega, \;\; \underset{  \Psi \bydef \{\veta \;|\; \|\veta\|<\alpha\}}{\underbrace{\|\veta\| \le \alpha}}.
    \label{eq: linear OPAD}
\end{align}
Therefore, to analyze OPAD, we just need to understand the sets $\Omega$, $\Psi$, and the parameter $\vtheta$.

\subsection{Geometry of the constraints}
There are two constraints in \eref{eq: linear OPAD}. The first constraint $\veta \in \Psi$ is a simple $\alpha$-ball surrounding the input. It says that the perturbation in the camera space should be bounded.

\begin{figure}[h]
\centering
\includegraphics[width=\linewidth]{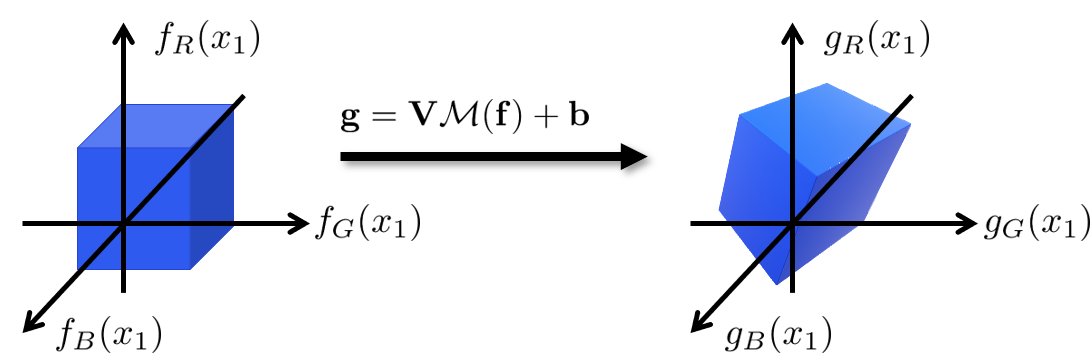}
\caption{The constraint $\Omega$ is constructed by converting a box constraint $0 \le \vf(x_1) \le 1$ to a rectangle via a per-pixel per-color transformation $\calM$, followed by a color mixing process. }
\label{fig:constraints}
\end{figure}

The more interesting constraint is $\veta \in \Omega$. To understand how $\Omega$ contributes to the feasibility of OPAD, we consider one pixel location $x_1$ of the object. This pixel has three colors $\vf(x_1) = [f_R(x_1), f_G(x_1), f_B(x_1)]^T$. Since $\vf(x_1)$ is the signal we send to the projector, it holds that $0 \le \vf(x_1) \le 1$ as illustrated in \fref{fig:constraints}.

After passing through the projector-camera model, the observed pixel by the camera is $\vg(x_1)$. The relationship between $\vg(x_1)$ and $\vf(x_1)$ is given by \eref{eq: T} (for pixel $x_1$). The conversion from $\vf(x_1)$ to $\vg(x_1)$ involves $\calM$ which is a pixel-wise and color-wise mapping. Since $\vf(x_1)$ lives in a unit cube, $\calM(\vf(x_1))$ will live in a rectangular box. The second conversion from $\calM(\vf(x_1))$ to $\vg(x_1)$ involves an affine transformation. Therefore, the resulting set that contains $\vg(x_1)$ will be a 3D polygon. This 3D polygon is $\Omega$.

\subsection{When will OPAD fail?}
\begin{figure}[ht]
\centering
\includegraphics[width=\linewidth]{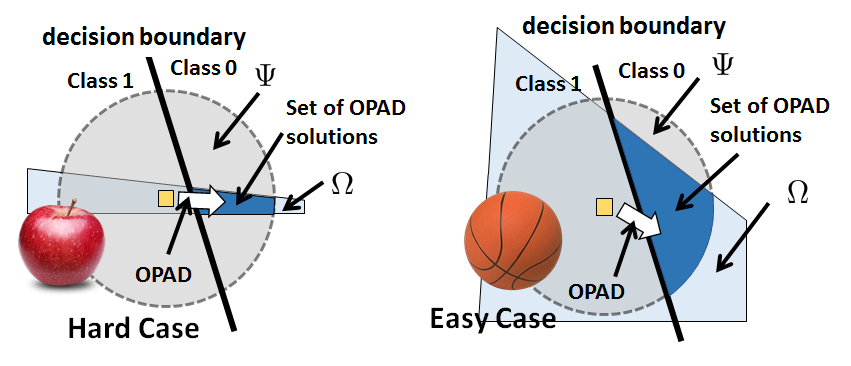}
\caption{(a) OPAD is hard when the color transformation $\mathbf{V}$ is nearly singular. This happens when the object has saturated pixels or is reflective. (b) OPAD is easy when $\Omega$ covers a large portion of the target class. This happens when $\mathbf{V}$ is invertible, e.g., for fabric, textile, rough surfaces etc.}
\label{fig:understandingV}
\vspace{-2ex}
\end{figure}

The feasibility of OPAD is determined by $\Omega \cap \Psi$ and the decision boundary $\vtheta$, as illustrated in \fref{fig:understandingV}. Given a clean classifier $\vtheta$, we partition the space into two half spaces (Class 1 and Class 0). The correct class is Class 1. To move the classification from Class 1 to Class 0, we must search along the feasible direction where $\Psi$ and $\Omega$ intersects.

\begin{figure*}[!ht]
\centering
\begin{tabular}{cccccccc}
True Obj. & \ \hspace{-2.0ex} Tgt apprnce. &\hspace{-2.0ex}Illumination &\hspace{-2.0ex}Captured  &\hspace{-2.0ex}True Obj. & \hspace{-2.0ex} Tgt apprnce. &\hspace{-2.0ex}Illumination &\hspace{-2.0ex}Captured  \\
\includegraphics[height=2.1cm]{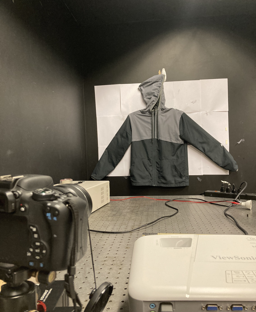}&
\hspace{-2.0ex}\includegraphics[height=2.1cm]{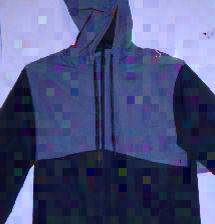}&
\hspace{-2.0ex}\includegraphics[height=2.1cm]{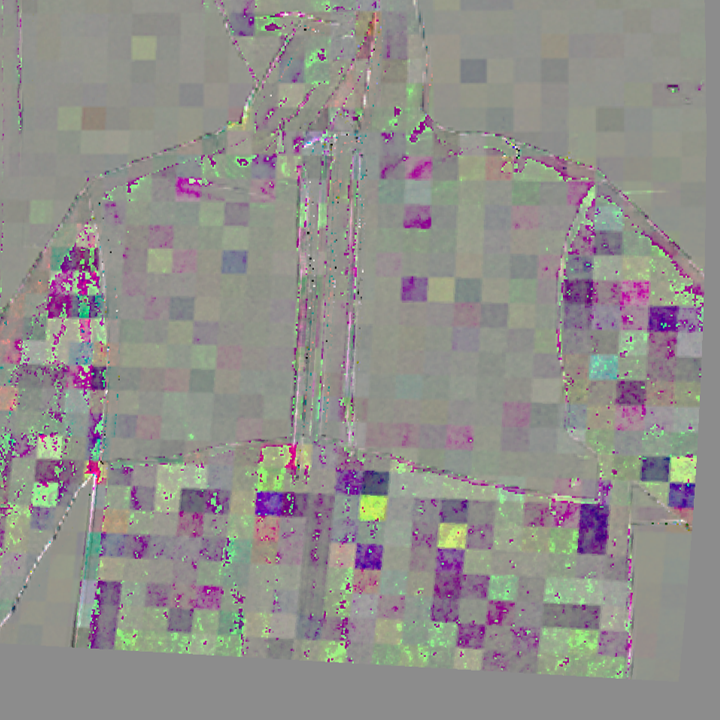}&
\hspace{-2.0ex}\includegraphics[height=2.1cm]{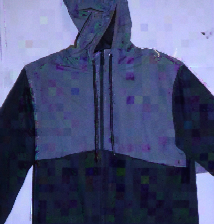}&
\hspace{-2.0ex}\includegraphics[height=2.1cm]{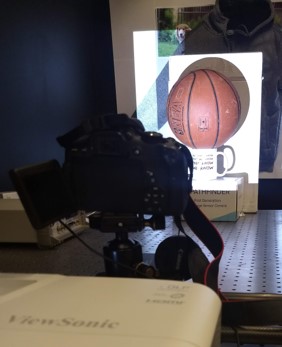}&
\hspace{-2.0ex}\includegraphics[height=2.1cm]{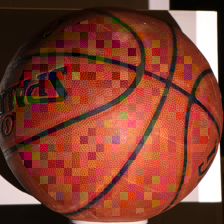}&
\hspace{-2.0ex}\includegraphics[height=2.1cm]{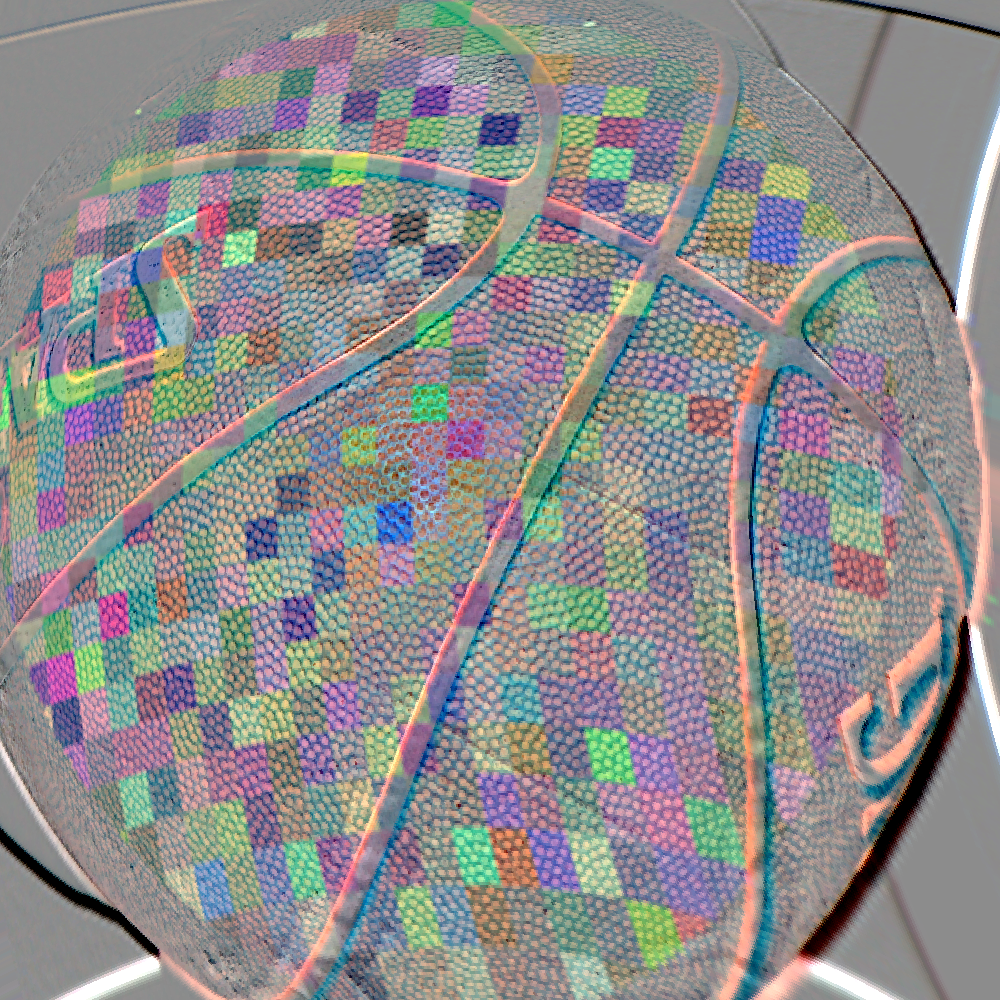}&
\hspace{-2.0ex}\includegraphics[height=2.1cm]{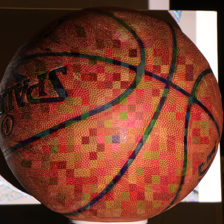}\\
Cardigan & \hspace{-2.0ex}Poncho   &\hspace{-2.0ex} &\hspace{-2.0ex} Poncho & \hspace{-2.0ex} Basketball &\hspace{-2.0ex} Buckler &\hspace{-2.0ex} &\hspace{-2.0ex} Buckler\\
\includegraphics[height=2.1cm]{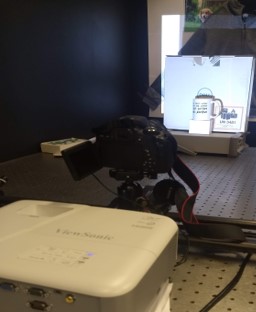}&
\hspace{-2.0ex}\includegraphics[height=2.1cm]{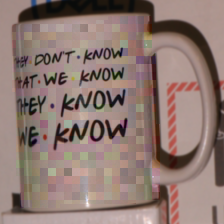}&
\hspace{-2.0ex}\includegraphics[height=2.1cm]{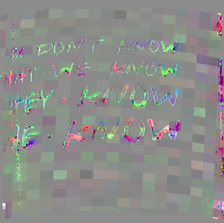}&
\hspace{-2.0ex}\includegraphics[height=2.1cm]{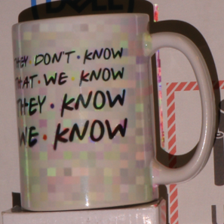}&
\hspace{-2.0ex} \includegraphics[height=2.1cm]{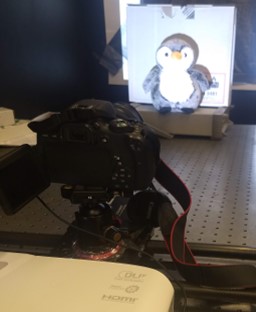}&
\hspace{-2.0ex}\includegraphics[height=2.1cm]{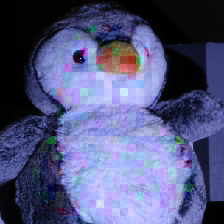}&
\hspace{-2.0ex}\includegraphics[height=2.1cm]{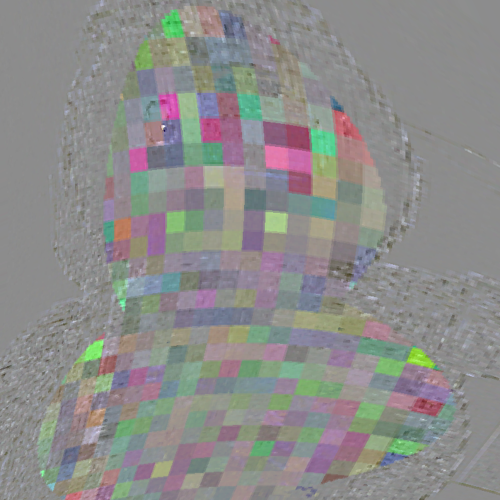}&
\hspace{-2.0ex}\includegraphics[height=2.1cm]{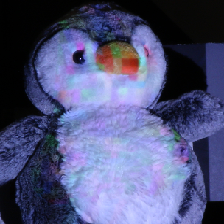}\\
Mug & \hspace{-2.0ex}Whiskey Jug   &\hspace{-2.0ex} &\hspace{-2.0ex} Whiskey Jug & Teddy & \hspace{-2.0ex}Wool   &\hspace{-2.0ex} &\hspace{-2.0ex} Wool
\end{tabular}

\vspace{1ex}
\begin{tabular}{ccccccc}
        \hline\hline
         Image & Algorithm & Generated on & Tested on & Block Size & Step-size& Norm. $\ell_2 $ Dist.\\
         \hline
         Cardigan   & PGD  ($\ell_\infty$)           & VGG-16 & VGG-16 & $8\times8$ &0.05 & 5.8/255\\
         Basketball & PGD ($\ell_\infty$)  & Resnet-50 & Resnet-50 &$8\times8$ &0.05&3.2/255 \\
         Coffee Mug & PGD ($\ell_2$)   & VGG-16 & VGG-16 &$8\times8$ &0.5&4.1/255 \\
         Teddy      & PGD ($\ell_2$)   & Resnet-50 & Resnet-50 &$8\times8$ &0.5 & 4.3/255 \\
         \hline
\end{tabular}

\vspace{1ex}
\caption{OPAD on real 3D objects. In each example, we show the targeted appearance (tgt apprnce) which is how the un-compensated attack should look like on a digital computer. The illumination is the OPAD illumination, and the capture is what the camera sees. Normalized $\ell_2$ distance measures the average $\ell_2$ difference between the original image and the captured image.}
\label{fig:read attacks}
\end{figure*}

The geometry above shows that the success/failure of OPAD is object dependent. Some objects are easier to attack, because the $\mathbf{V}$ matrix and the $\mathbf{b}$ vector create a ``bigger'' $\Omega$. This happens when the object surface is responsive to the illumination, e.g. the basketball shown in \fref{fig:understandingV}. The hard cases happen $\mathbf{V}$ and $\mathbf{b}$ create a very ``narrow'' $\Omega$. For example, a bright red color shirt is difficult because its red pixel is too strong. An apple is difficult because it reflects the light. Note that this is an intrinsic problem of the optics, and not the problem of the algorithm.

\subsection{Can we make OPAD unnoticeable?} \label{sec:unnotice}
The short answer is no. Unlike digital attacks where the average $\ell_2$ (or $\ell_\infty$) distance is driven by the decision boundary, the minimum amount of perturbation in OPAD is driven by the decision boundary \emph{and} the optics. The perturbation has to go through the radiometric response of the projector and the spectral response of the scene, not to mention other optical limits such as diffraction and out-of-focus. For tough surfaces, if the feasible set $\Omega$ is small, we have no choice but to increase the perturbation strength.

The conclusion of OPAD may appear pessimistic, because there are many objects we cannot attack. However, on the positive side OPAD suggests ways to \emph{defend} optical attacks. For example, one can configure the environment such that certain key features of the object are close to being saturated. Constantly illuminating the object with pre-defined patterns will also help defending against attacks. These are interesting topics for future research.

\section{Experiments}
We report experimental results on real 3D objects. For the results in the main paper, we mainly use white-box projected gradient descents (PGD) \cite{madry2017towards}, and fast gradient sign method (FGSM) \cite{goodfellow_14_explaining_adversarial}  to attack the objects. Additional results using black-box and other attack algorithms (e.g. colorization \cite{bhattad_19_color_att}) are reported in the supplementary material.

\subsection{Quantitative evaluation}
\label{sec: quant}
We first conduct a quantitative experiment on four real 3D objects (teddy, cardigan, basketball, and mug) as illustrated in \fref{fig:read attacks}. For each object, we generate 16 different targeted attacks: 4 different target classes (poncho, buckler, whiskey jug and wool), 2 different constraints ($\ell_2$ and $\ell_\infty$), and 2 different classifiers (VGG-16, and Resnet-50). The PGD \cite{madry2017towards} is used for all the 64 attacks. The parameter $\alpha$ was set to be $0.05$ for $\ell_\infty$ constraints and $0.5$ for $\ell_2$ constraints. We use the same gradient for each $8\times8$ pixels. 20 iterations were used for generating each attack.

The result in Table~\ref{fig:quant1} indicates that OPAD worked for 31 times out of the total 64 attacks (48\%). While this may not appear as a high success rate, the result is valid. The reason is that the success rate depends the specific types of objects being attacked. For example, the cardigan and the ball are easier to attack, but the teddy and the mug are difficult to attack. This is consistent with our theoretical analysis in \fref{fig:understandingV}, where certain objects have a small feasible set $\Omega$ due to the poor spectral response. We emphasize that this is the limitation of the optics, and not the attack optimization.

\begin{table}[h]
\centering
\begin{tabular}[t]{ccccc}
\hline
\hline
Tar./Obj.&Cardigan&Teddy&Ball&Mug  \\
\hline
Poncho&4/4, 0.85 &0/4, 0.13&2/4, 0.46&0/4, 0.10\\
Wool&4/4, 0.74&3/4, 0.22&1/4, 0.27&0/4, 0.14\\
Buckler&4/4, 0.48&0/4, 0.08&3/4, 0.72& 1/4, 0.18\\
Jug&4/4, 0.65&0/4, 0.05&2/4, 0.51&3/4, 0.34\\
\hline
\end{tabular}
\vspace{1ex}
\caption{ Success rate and confidence for the targeted attack experiments. Notice that the success of the attack depends on the object being attacked. While Cardigan and ball are easier to attack, teddy and the mug are not.}
\label{fig:quant1}
\vspace{-1ex}
\end{table}

In \tref{fig:quant2}, we compare the influence due to the classification network and perturbation $\ell_p$-norm constraints. The results indicate that VGG-16 is easier to attack than ResNet-50, and $\ell_2$ is slightly easier to attack than $\ell_\infty$. However, the difference is not significant. We believe that the optics plays a bigger role here.

\begin{table}[h]
\centering
\begin{tabular}[t]{cc|cc}
\hline
\hline
\multicolumn{2}{c}{Network} & \multicolumn{2}{c}{Constraint}\\
\hline
VGG-16 & Resnet-50 & $\ell_2$ & $\ell_\infty$ \\
17/32, 0.43 & 14/32, 0.31 & 16/32, 0.40 & 15/32, 0.34\\
\hline
\end{tabular}
\vspace{1ex}
\caption{Success rate and confidence for targeted attack using different networks and different constraints.}
\label{fig:quant2}
\vspace{-1ex}
\end{table}

\subsection{Need for projector-camera compensation} \label{sec:comp}
This experiment aims to verify the necessity of the projector-camera compensation step in OPAD. To reach a conclusion, we consider another projector-based attack method proposed by Nguyen et al. \cite{CVPRW_attack}. It is a \emph{single-capture} optical attack like our proposed method. Its projector-camera compensation consists of a simple global color correction without taking the spatially varying color mixing matrix $\mathbf{V}$ into consideration.

We attack a VGG-16 network using PGD, with $\alpha = 1$. Both the algorithms were run for 20 iterations. We attacked the object `book' by targeting 15 random classes from the ImageNet dataset \cite{imagenet_cvpr09}.  Table~\ref{table: compensation} summarizes the quantitative evaluation results (with details in the supplementary materials). OPAD worked 10/15 times, while \cite{CVPRW_attack} worked only 3/15 times.

\begin{table}[h]
\scalebox{0.9}{
\begin{tabular}{cccccccc}
\hline
\hline
\multicolumn{8}{c}{From ``book'' to one of the 15 target classes}\\
\hline
 1 & 2 & 3 & 4 & 5 & 6 & 7 & 8 \\
 $\checkmark/\times$ & $\checkmark/\times$ & $\times/\times$
& $\checkmark/\checkmark$ & $\checkmark/\times$ & $\checkmark/\times$ & $\checkmark/\times$ & $\times/\times$  \\
\hline
 9 & 10 & 11 & 12 & 13 & 14 & 15 &  \\
 $\checkmark/\checkmark$ & $\times/\times$ & $\checkmark/\times$
& $\times/\times$ & $\times/\times$ & $\checkmark/\checkmark$ & $\checkmark/\times$   \\
\hline
\end{tabular}
}
\vspace{1ex}
\caption{Comparing (with OPAD / with \cite{CVPRW_attack}) when attacking a real 3D book to 15 target classes. Out of the 15 attacks, an attack with OPAD succeeds in 10/15 time  whereas an attack without OPAD succeeds in only 3/15 times.}
\label{table: compensation}
\end{table}

\subsection{How strong should OPAD be?}
\begin{figure}[ht]
\centering
\begin{tabular}{ccccc}
\hspace{-2.0ex} Target&
\hspace{-2.0ex} $\alpha = 0.1$ &
\hspace{-2.0ex} $\alpha = 0.5$ &
\hspace{-2.0ex} $\alpha = 1.0$ &
\hspace{-2.0ex} $\alpha = 1.5$ \\
\hspace{-2.0ex} \includegraphics[width=0.19\linewidth]{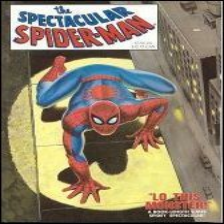} &
\hspace{-2.5ex} \includegraphics[width = 0.19\linewidth]{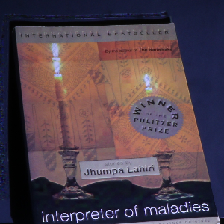} &
\hspace{-2.5ex} \includegraphics[width = 0.19\linewidth]{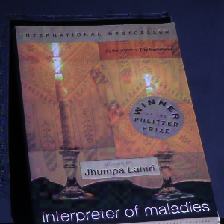} &
\hspace{-2.5ex} \includegraphics[width = 0.19\linewidth]{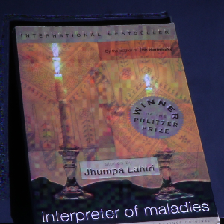} &
\hspace{-2.5ex} \includegraphics[width = 0.19\linewidth]{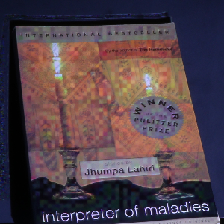} \\
\hspace{-2.0ex} Comics &
\hspace{-2.0ex} Book &
\hspace{-2.0ex} Packet &
\hspace{-2.0ex} Comics &
\hspace{-2.0ex} Comics \\
\hspace{-2.0ex} \includegraphics[width=0.19\linewidth]{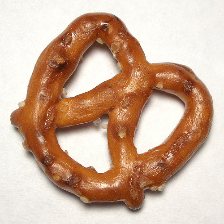} &
\hspace{-2.5ex} \includegraphics[width = 0.19\linewidth]{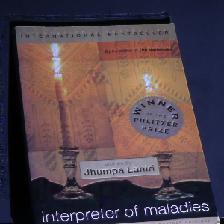} &
\hspace{-2.5ex} \includegraphics[width = 0.19\linewidth]{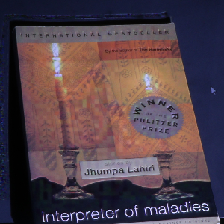} &
\hspace{-2.5ex} \includegraphics[width = 0.19\linewidth]{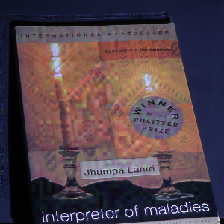} &
\hspace{-2.5ex} \includegraphics[width = 0.19\linewidth]{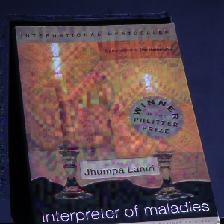} \\
\hspace{-2.0ex} Pretzel &
\hspace{-2.0ex} Book &
\hspace{-2.0ex} Book &
\hspace{-2.0ex} Packet &
\hspace{-2.0ex} Pretzel\\
\end{tabular}
\vspace{1ex}
\caption{Unlike digital attacks where the perturbation is determined by the $\ell_p$-ball, OPAD needs to overcome the optics. The experiment here shows the amount of perturbation required to turn a ``book'' to a ``comic-book'' and a ``pretzel''.}
\label{fig:imperceptibility}
\vspace{-2ex}
\end{figure}

A natural question following Table~\ref{table: compensation} is the minimum perturbation strength that is needed for OPAD. As discussed in Section 4.3, OPAD is fundamentally limited by the optics and the decision boundary. Unlike digital attacks where the $\ell_p$ ball can be made very small, OPAD attack has to be reasonably strong to compensate for the optical loss. In \fref{fig:imperceptibility}, we conduct an experiment to understand how imperceptible OPAD can be. We want to turn a ``book'' into a  ``comic-book'' or a ``pretzel''. For the both targets, we launch 4 attacks using $\alpha \in \{0.1, 0.5, 1.0, 1.5\}$. We see that a smaller $\alpha$ is sufficient for ``comic-book'' and a larger $\alpha$ is needed for ``pretzel''. In both cases, the perturbation is not too strong but still visible.

\begin{figure*}[!ht]
\centering
\begin{tabular}{ccccc}
True Object & \hspace{-2.0ex} w/o $\Omega$ (on computer)& \hspace{-2.0ex} w/ $\Omega$ (on computer) &\hspace{-2.0ex}Illumination &\hspace{-2.0ex}Captured (on camera) \\
\includegraphics[width=0.175\linewidth]{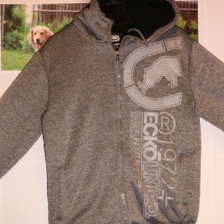}&
\hspace{-2.0ex}\includegraphics[width=0.175\linewidth]{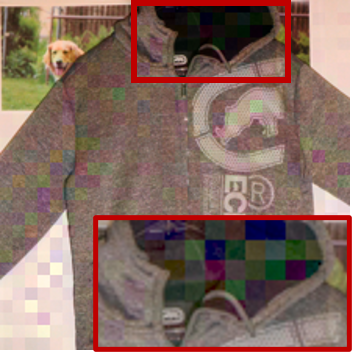}&
\hspace{-2.0ex}\includegraphics[width=0.175\linewidth]{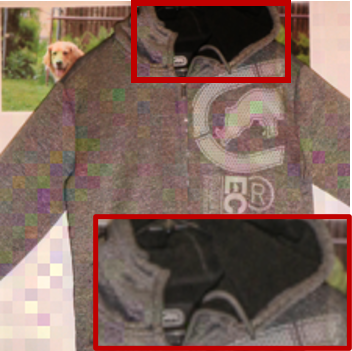}&
\hspace{-2.0ex}\includegraphics[width=0.175\linewidth]{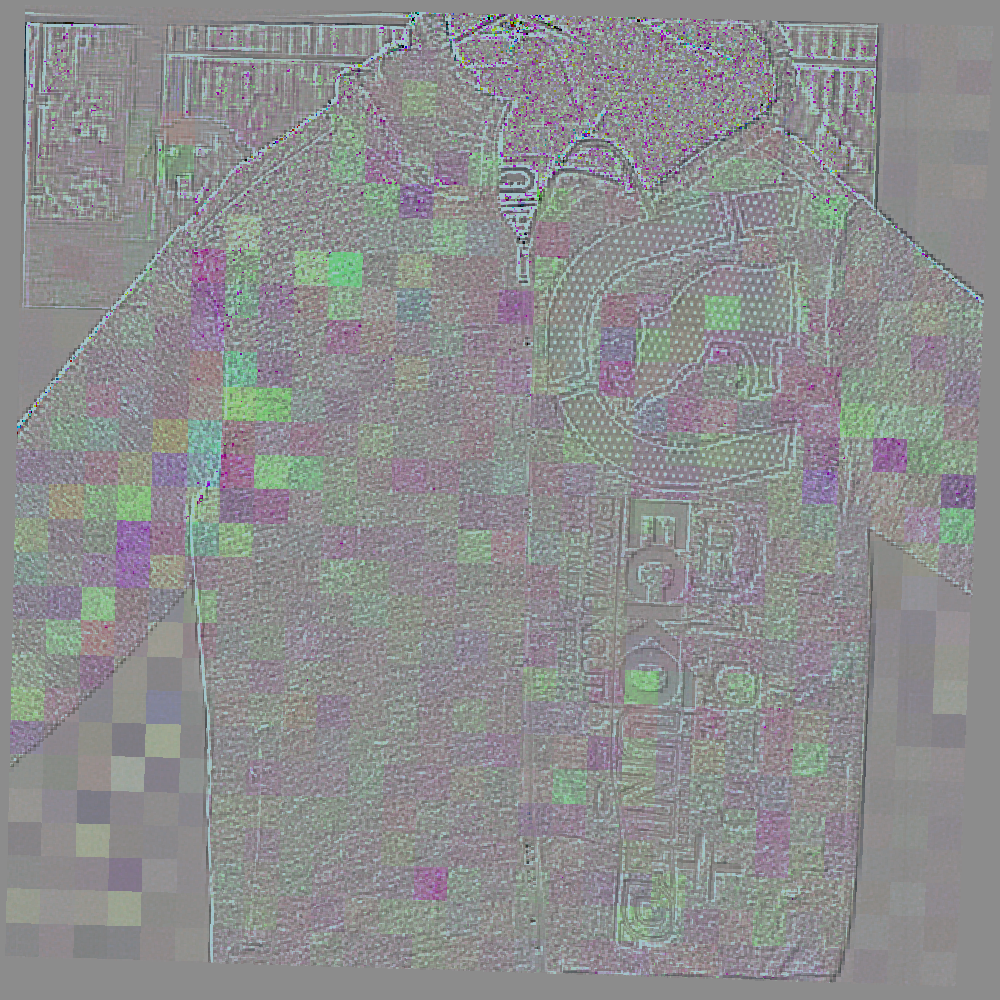}&
\hspace{-2.0ex}\includegraphics[width=0.175\linewidth]{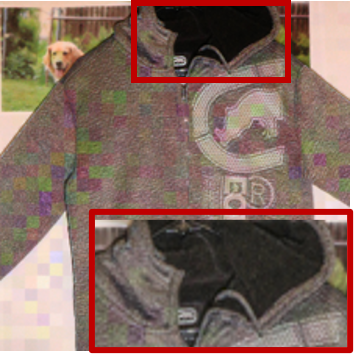}\\
Cardigan & Vestment & Sweatshirt & & Sweatshirt
\end{tabular}
\caption{Significance of the constraint $\veta \in \Omega$. Notice that running the optimization without $\Omega$, we will generate an image that may not be optically achievable. The inset images are displayed with MATLAB's tonemap function. }
\label{fig:Jacket_proposedPGD}
\end{figure*}

\begin{figure*}[!ht]
\centering
\begin{tabular}{ccccc}
True Obj. & \ \hspace{-2.0ex} OPAD &\hspace{-2.0ex} Translation &\hspace{-2.0ex} Zoom &\hspace{-2.0ex} Failure Case \\
\includegraphics[width=0.18\linewidth]{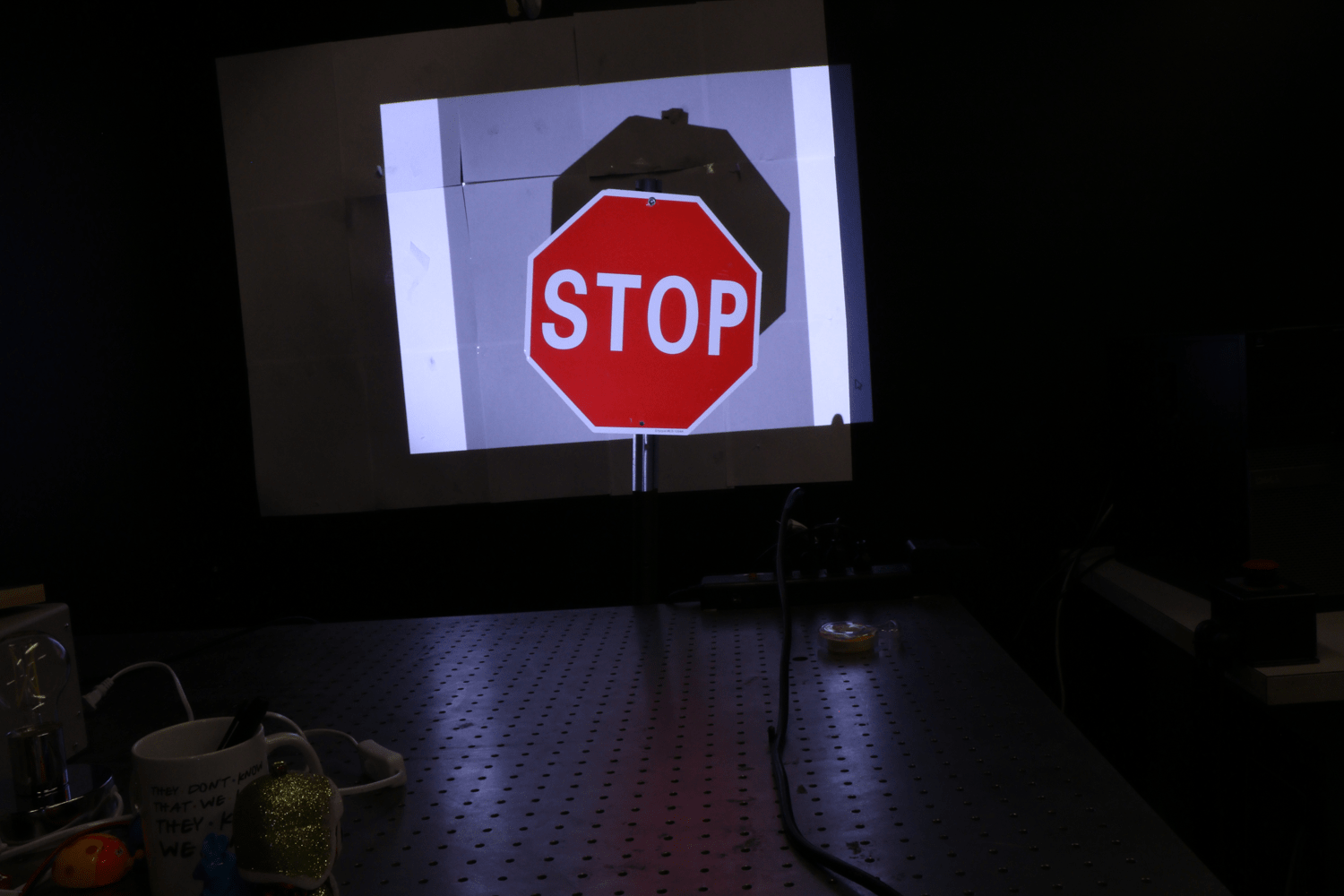}&
\hspace{-2.0ex}\includegraphics[width=0.18\linewidth]{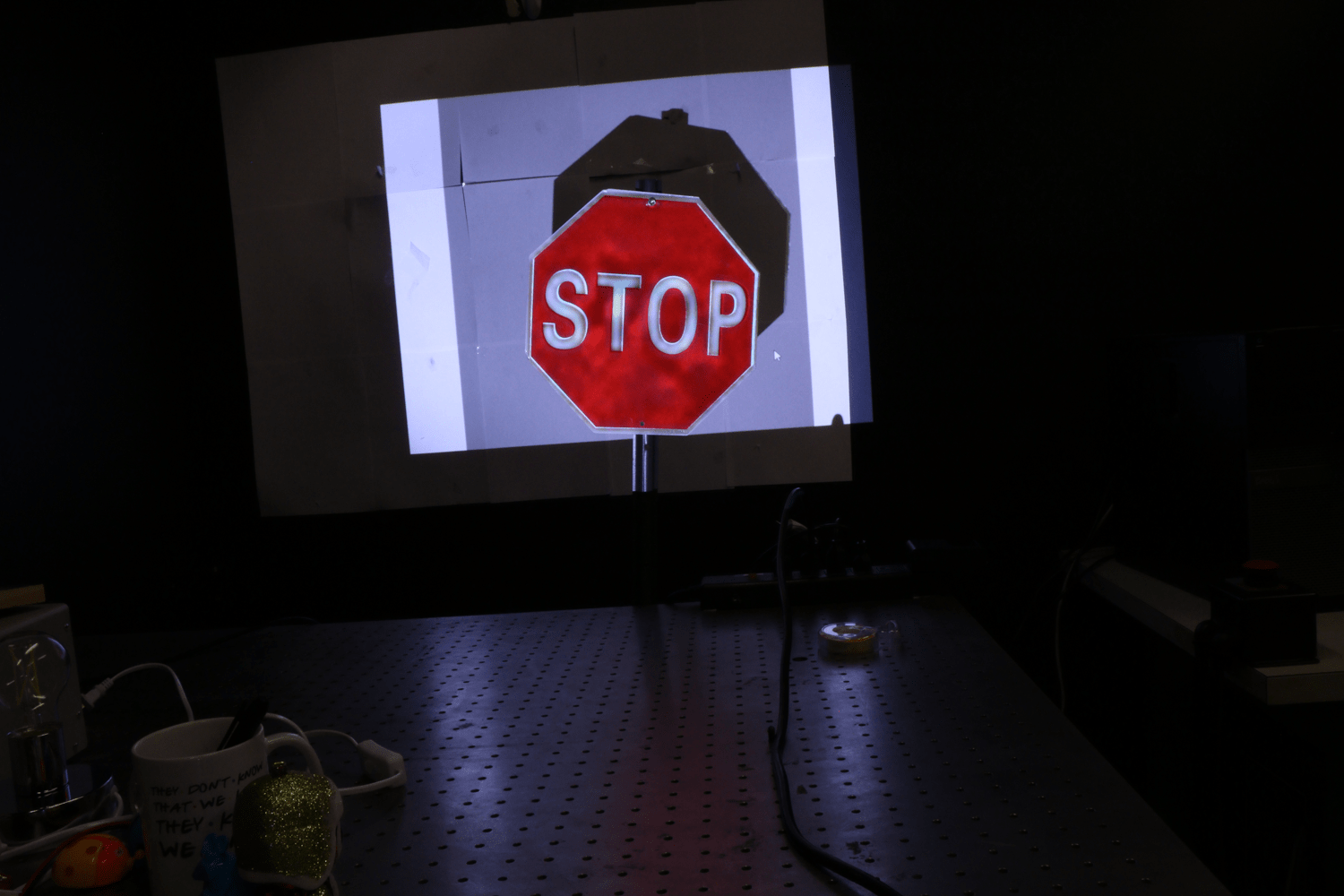}&
\hspace{-2.0ex}\includegraphics[width=0.18\linewidth]{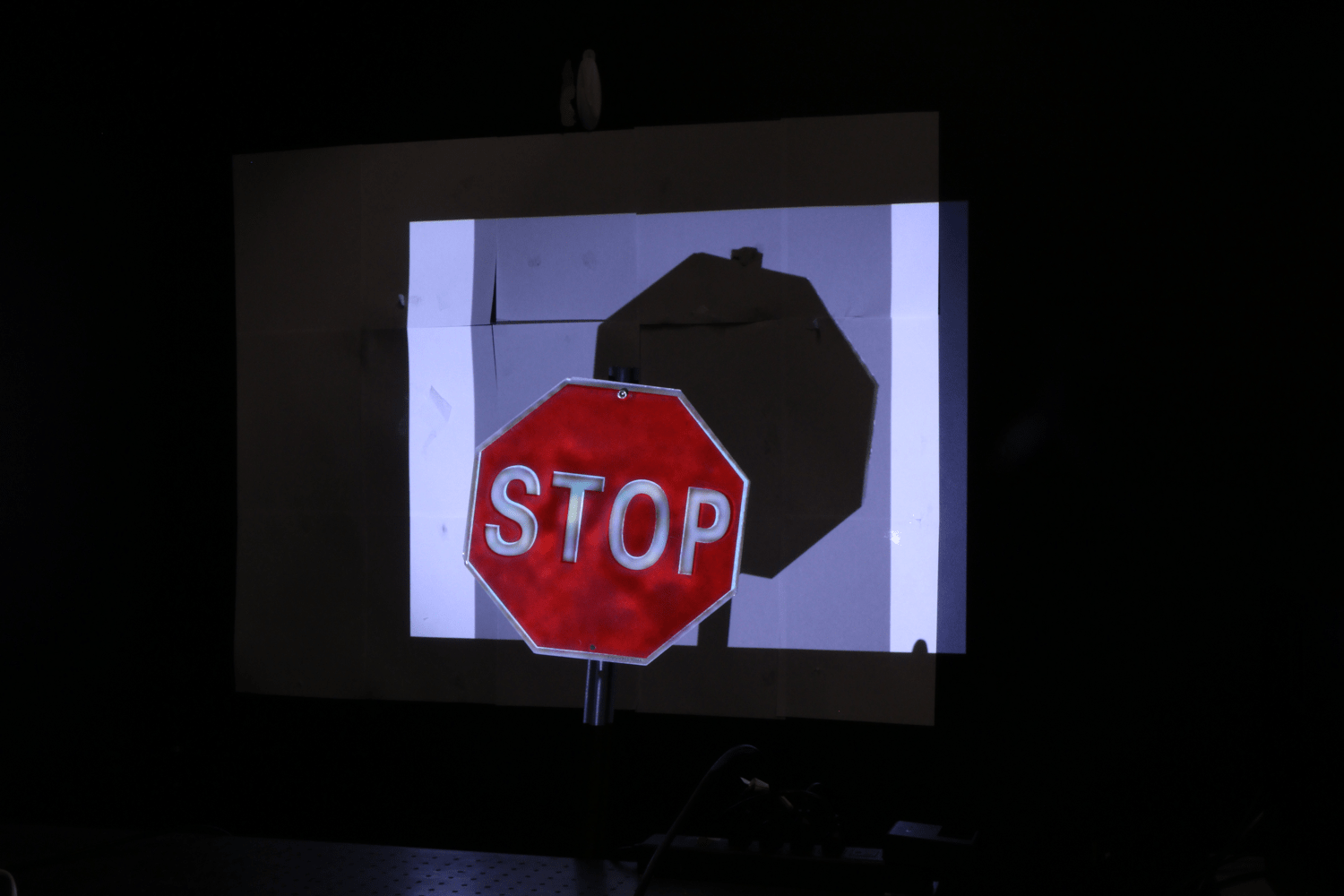}&
\hspace{-2.0ex}\includegraphics[width=0.18\linewidth]{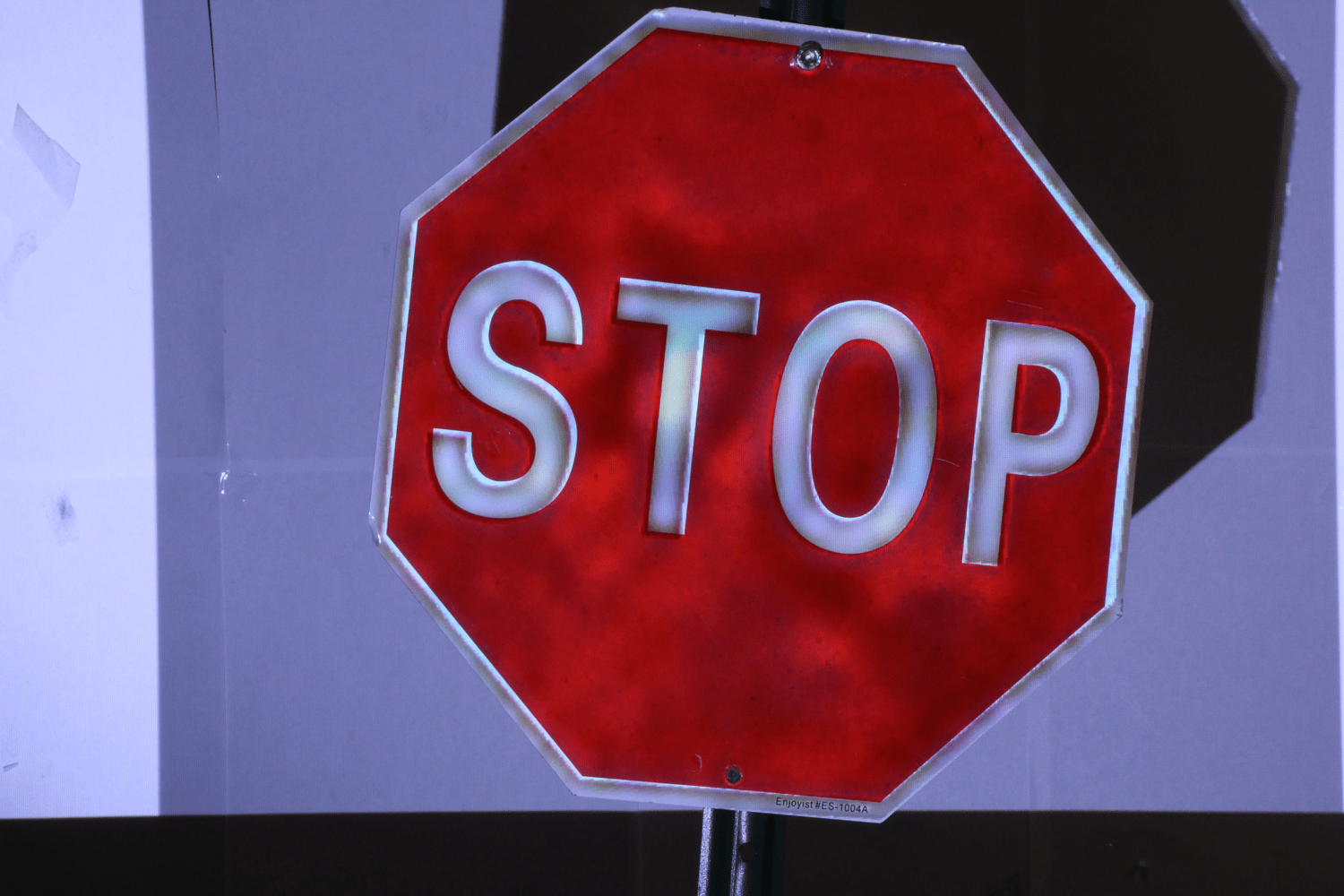}&
\hspace{-2.0ex}\includegraphics[width=0.18\linewidth]{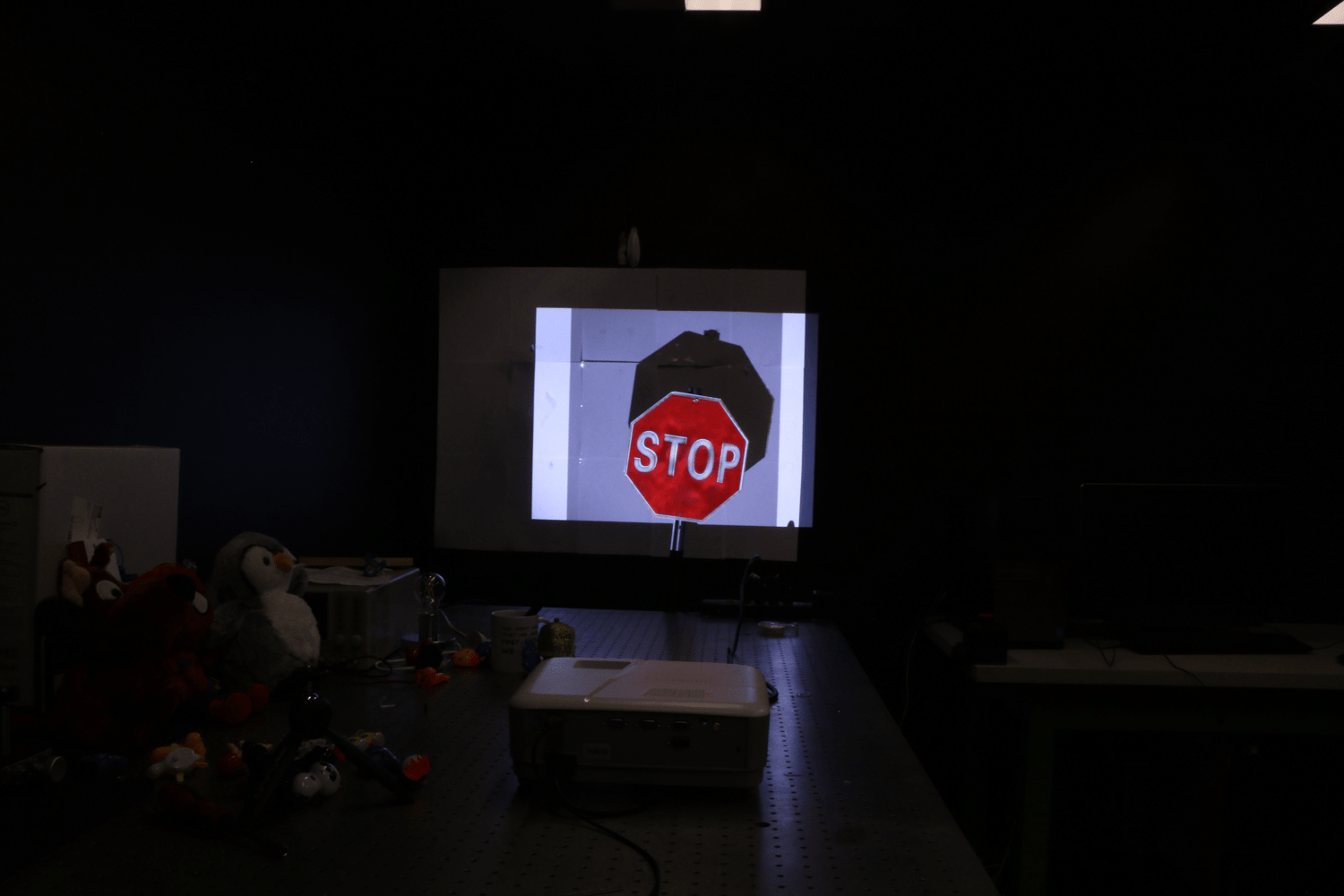}\\
Stop Sign & \hspace{-2.0ex}Speed 60   &\hspace{-2.0ex} Speed 60 &\hspace{-2.0ex} Speed 60 & \hspace{-2.0ex} Stop Sign
\end{tabular}
\caption{Robustness of OPAD against perspective change (using a real metallic STOP sign). As we translate the camera, the same OPAD illumination is still capable of attacking. The failure happens when the camera zooms out too much.}
\label{fig:Ablation_STOP}
\end{figure*}

\begin{figure*}[!ht]
\centering
\begin{tabular}{cccccc}
True, ISO 800 & \hspace{-2.0ex} Att. ISO 200 &\hspace{-2.0ex} Att. ISO 400 &\hspace{-2.0ex} Att. ISO 800 &\hspace{-2.0ex} Att. ISO 1600 & \hspace{-2.0ex} Att. ISO 3200 \\
\includegraphics[width=0.14\linewidth]{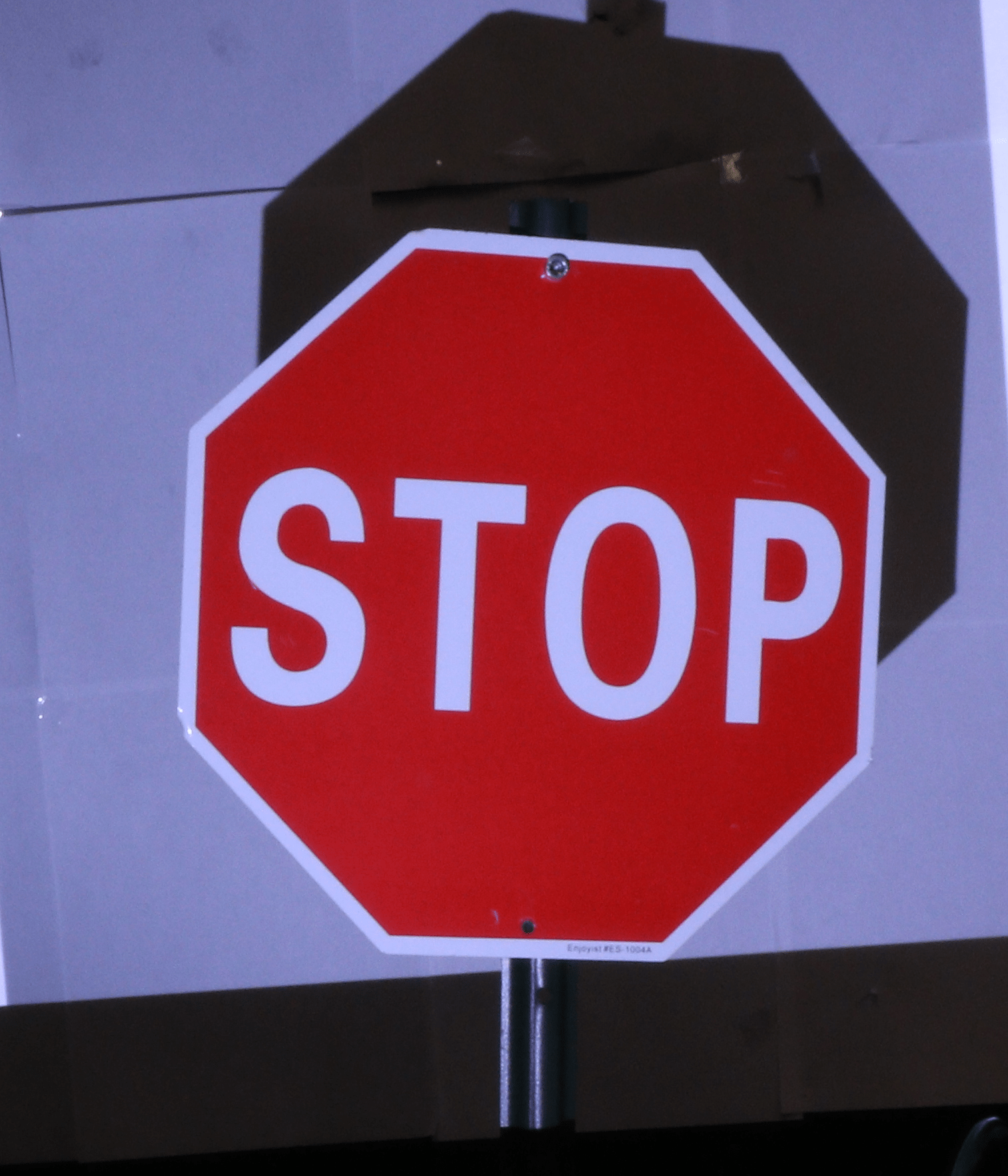}&
\hspace{-2.0ex}\includegraphics[width=0.14\linewidth]{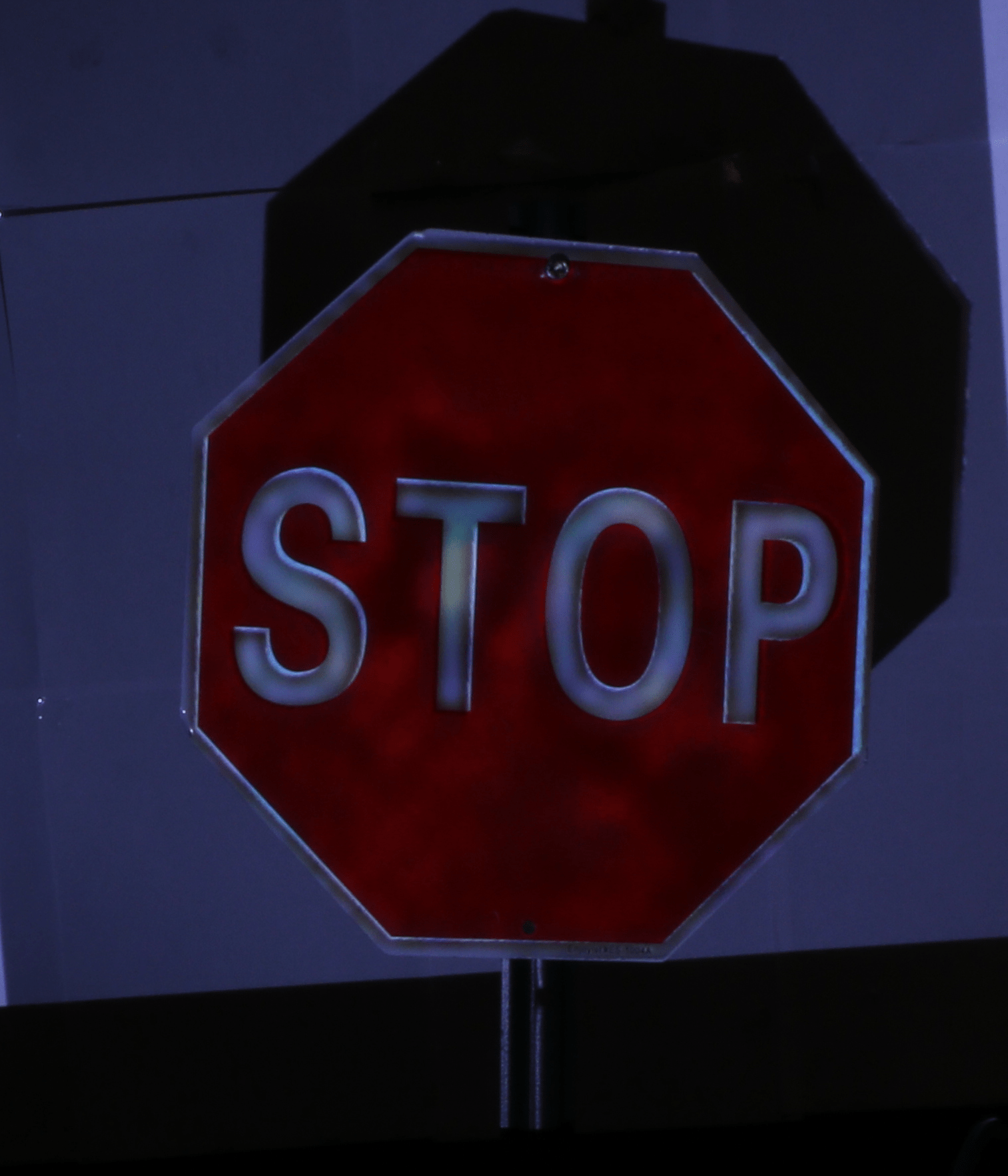}&
\hspace{-2.0ex}\includegraphics[width=0.14\linewidth]{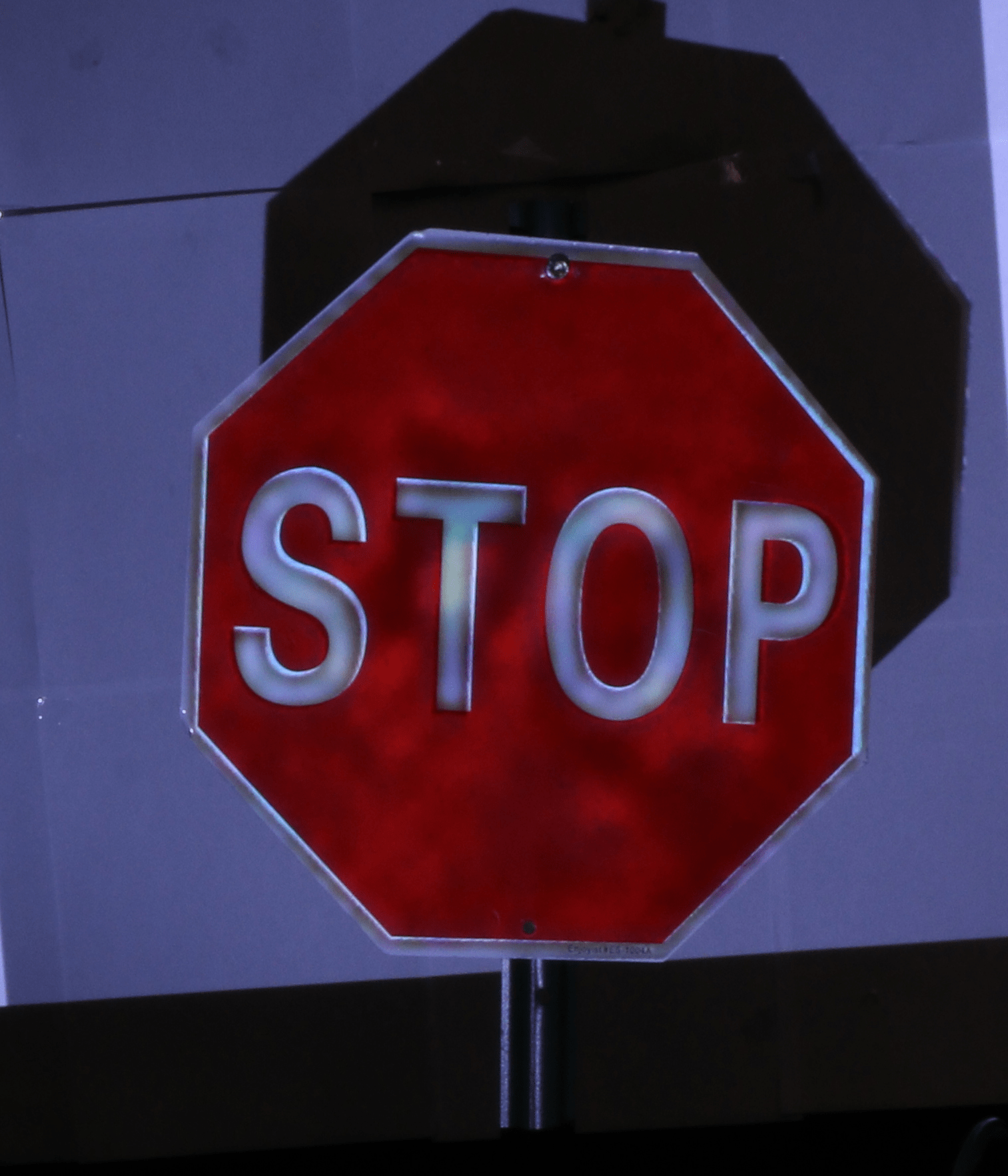}&
\hspace{-2.0ex}\includegraphics[width=0.14\linewidth]{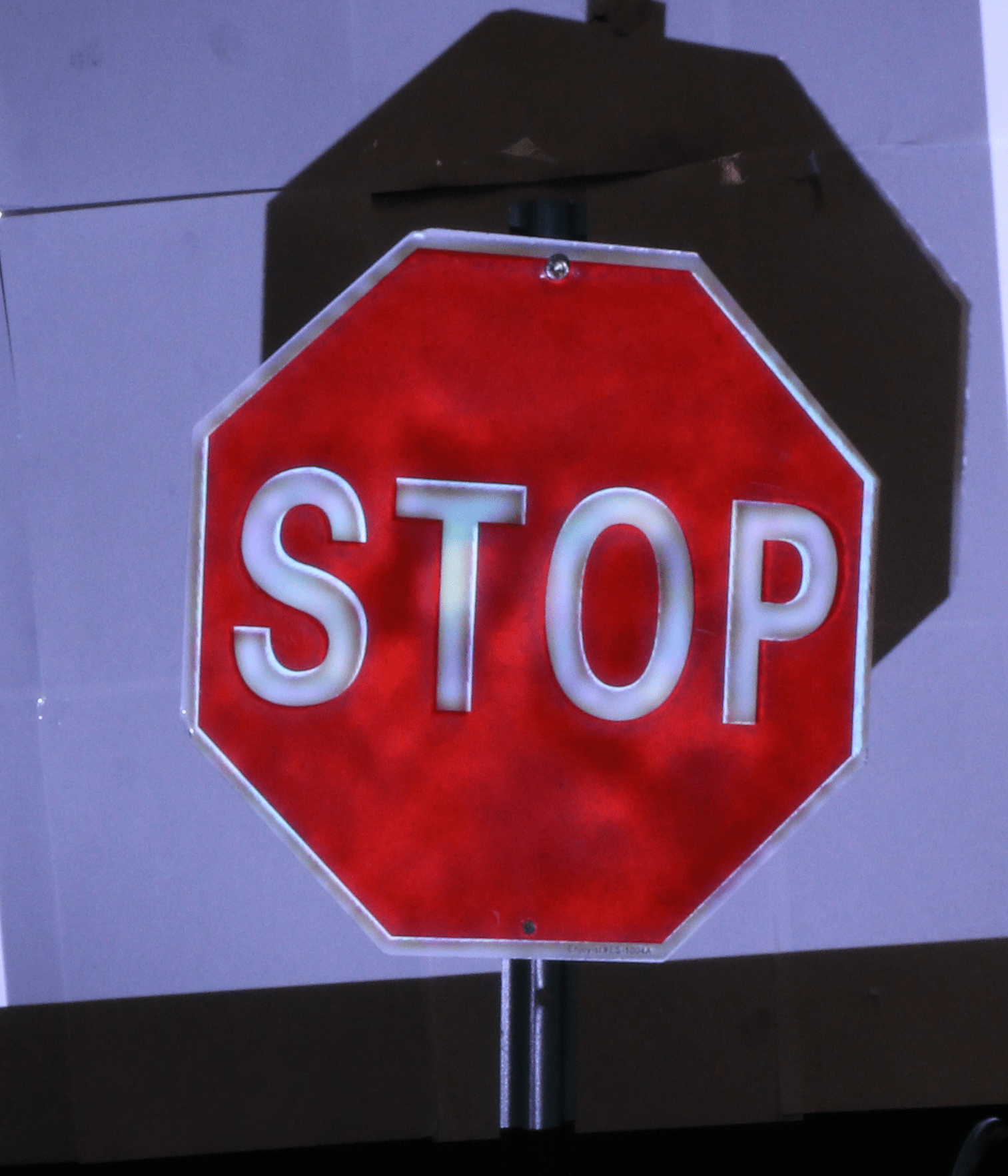}&
\hspace{-2.0ex}\includegraphics[width=0.14\linewidth]{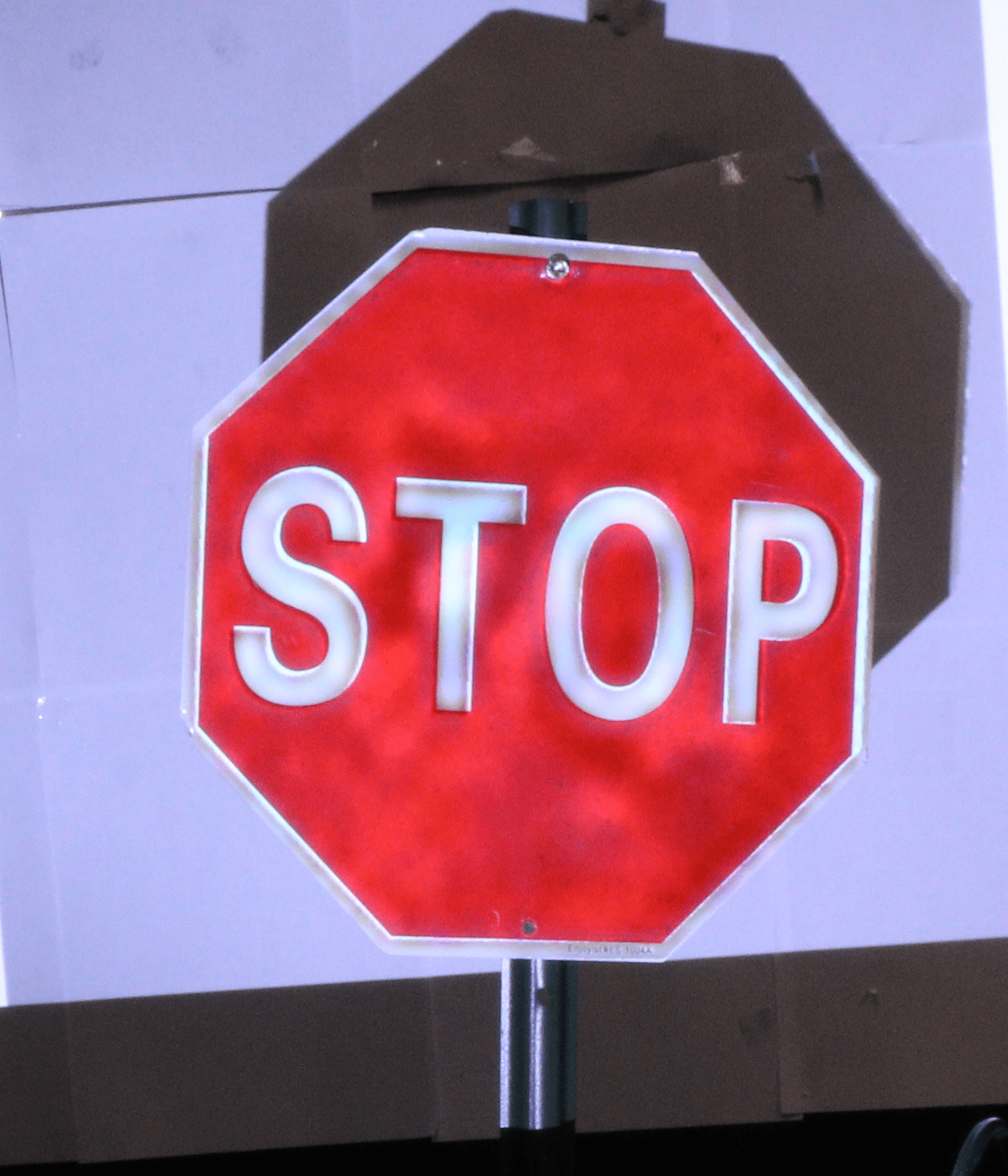}&
\hspace{-2.0ex}\includegraphics[width=0.14\linewidth]{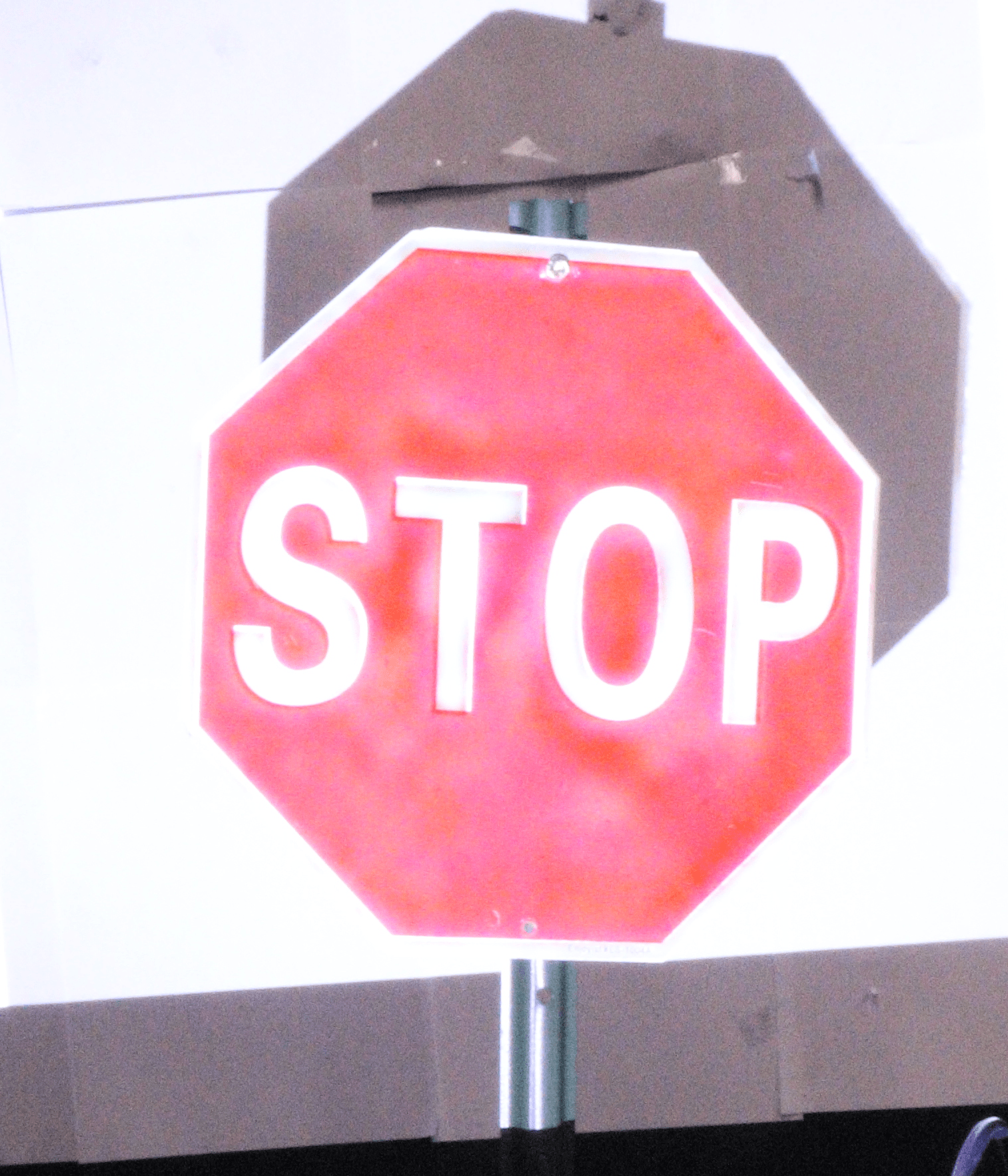}\\
Stop Sign & \hspace{-2.0ex}Speed 60   &\hspace{-2.0ex} Speed 60 &\hspace{-2.0ex} Speed 60 & \hspace{-2.0ex}Speed 60 & \hspace{-2.0ex} Stop Sign
\end{tabular}
\caption{Robustness of OPAD against different ISO of the camera (using a real metallic STOP sign). The OPAD attack is computed for ISO 800, but the images are captured at other ISO levels. The failure happens when the ISO is too high so that many pixels are saturated. }
\label{fig:Ablation_STOP_ISO}
\end{figure*}

\subsection{Significance of the constraint $\Omega$} \label{sec:veta}
In this experiment we shift our attention to the constraint $\Omega$ because it is this constraint $\Omega$ that makes our problem special. We ask: \textit{what happens if we ignore the constraint $\veta \in \Omega$ in the optimization?} The short answer is that we will generate illumination patterns that are not feasible. To justify this claim, we conduct an experiment by launching a white-box FGSM attack \footnote{We can also attack using PGD but a single step attack will demonstrate the significance of $\Omega$ more clearly.} on VGG-16 for a real 3D cardigan shown in \fref{fig:Jacket_proposedPGD}. The result shows that if we ignore the constraint, FGSM will generate a pattern that contains colors that are not achievable. In contrast, when $\Omega$ is included, the optimization solution will be optically feasible.

\subsection{Robustness against perspective and ISO}\label{sec:robustness}
Our final experiment concerns about the robustness of OPAD against translation, zoom, and varying ISO settings. This is important because OPAD can potentially be used to fool a neighboring camera and not just the OPAD camera. We conduct two experiments using a real metallic STOP sign. The classifier is trained based on \cite{eykholt_18_stopsign_phy}, using German Traffic Sign Recognition Benchmark dataset \cite{stallkamp2012man}.

In \fref{fig:Ablation_STOP}, we capture the scene with different camera location and zooms. We first generate a successful attack on the STOP sign, which is classified as `Speed limit 60'. The camera is then translated by $30^\circ$ w.r.t the position of the STOP sign. We also capture the scene with zoom in and zoom out. The result shows that the OPAD still works until the object is zoomed out for a long distance.

In \fref{fig:Ablation_STOP_ISO}, we adjust the camera with different ISO settings. The attack is generated using an ISO setting of 800. As the ISO changes in the range $[200,400,800,1600,3200]$, OPAD remains robust except for 3200 ISO where a lot of pixels are saturated.

\section{Discussions and conclusion}
\textbf{Broader impact of OPAD}. While this paper focuses exclusively on demonstrating how to \emph{attack}, OPAD has the potential to address the critical need in robust machine learning today where we do not have a way to model the environment. OPAD provides a parametric model where the parameters are controlled through the hardware. If we want to mimic an environment, we can adjust the OPAD parameters until the scene is reproduced (or report that it is infeasible). Consequently, we can analyze the robustness of the classifiers and defense techniques. We note that none of the existing optical attacks has this potential.

\textbf{Conclusion}. OPAD is a non-invasive adversarial attack based on structured illumination. For a variety of existing attack methodologies (targeted, untargeted, white-box, black-box, FGSM, PGD, and colorization), OPAD can transform the known digital results into real 3D objects. The feasibility of OPAD is constrained by the surface material of the object and the saturation of color. The success of OPAD demonstrates the possibility of using an optical system to alter faces or for long-range surveillance tasks. It would be interesting to see how these can be realized in the future.

\section{Acknowledgement}
The work is funded partially by the Army Research Office under the contract W911NF-20-1-0179 and the National Science Foundation under grant ECCS-2030570.

\section{Supplementary document}
\section*{Estimating the model parameters}
In this section we discuss how to estimate the forward model $\calT(\mathbf{f})$ and the the inverse function $\calT^{-1}(\mathbf{g})$. Estimating the inverse function follows directly from \cite{grossberg_04_ShreeNayar_calibration}. We then extend it to obtain the forward model too. 

\subsection*{Notations}
We denote the input to the projector at the $x$-th pixel as $\vf(x) = [f_R(x), f_g(x), f_B(x)]^T$. The corresponding pixel in the captured image is $\vg(x)$. The radiometric response function of the projector is denoted by 
\begin{equation*}
\vz(x) \bydef 
\begin{bmatrix}
z_R(x)\\
z_G(x)\\
z_B(x)
\end{bmatrix}
= 
\begin{bmatrix}
\calM_R(f_R(x))\\
\calM_G(f_G(x))\\
\calM_B(f_B(x))
\end{bmatrix}
= \calM(\vf(x)).
\end{equation*}

As outlined in the main paper, the relation between the projector input and the captured image at any pixel can be written as
\begin{equation}
    \vg(x) = \mV^{(x)} \vz(x) + \vb^{(x)}
\end{equation}
where $\mV^{(x)}$ is the color mixing matrix defined as
\begin{equation}
    \mV^{(x)}=                                      
    \begin{bmatrix}
V_{RR}^{(x)} & V_{RG}^{(x)} & V_{RB}^{(x)}\\
V_{GR}^{(x)} & V_{GG}^{(x)} & V_{GB}^{(x)}\\
V_{BR}^{(x)} & V_{BG}^{(x)} & V_{BB}^{(x)}
\end{bmatrix},
\end{equation}
and $\vb^{(x)}$ is the offset.

We also define another matrix $\widetilde{\mV}^{(x)} \bydef \mV^{(x)} \mD^{(x)^{-1}}$, where  $\mD^{(x)} \bydef \text{diag}\left\{V_{RR}^{(x)}, V_{GG}^{(x)}, V_{BB}^{(x)}\right\}$. So, $\mV^{(x)} = \widetilde{\mV}^{(x)} \mD^{(x)}$. The utility of this matrix will become clear in later part of this document. 

For notation simplicity, in the rest of this supplementary material we drop the coordinate $x$ because all equations are pixel-wise.

\subsection*{Estimating $\widetilde{\mV}^{(x)}$}

Consider two images obtained by changing only the red channel input between them. These two inputs can be written as $\vf^{(1)} = [f_R^{(1)}, f_G, f_B] $, and $\vf^{(2)} = [f_R^{(2)}, f_G, f_B]$. Following the projector model, the two captured pixels $\vg^{(1)}$ and $\vg^{(2)}$ can be written as 
\begin{align*}
\begin{bmatrix}
g_{R}^{(1)}\\
g_{G}^{(1)}\\
g_{B}^{(1)} 
\end{bmatrix} &= \widetilde{\mV} \mD 
\begin{bmatrix}
{z_R^{(1)}}\\
z_G\\
z_B 
\end{bmatrix} + \vb,\\
\begin{bmatrix}
g_{R}^{(2)}\\
g_{G}^{(2)}\\
g_{B}^{(2)} 
\end{bmatrix} &= \widetilde{\mV} \mD 
\begin{bmatrix}
{z_R^{(2)}}\\
z_G\\
z_B 
\end{bmatrix} + \vb
\end{align*}
By taking the difference between the two images we obtain,

\begin{align}\label{eqn:Deltas}
\begin{bmatrix}
\Delta g_R\\
\Delta g_G\\
\Delta g_B
\end{bmatrix}
&\bydef
\begin{bmatrix}
g_R^{(1)} - g_R^{(2)}\\
g_G^{(1)} - g_G^{(2)}\\
g_B^{(1)} - g_B^{(2)}
\end{bmatrix} \notag \\
&=
\begin{bmatrix}
\widetilde{V}_{RR} V_{RR} \left(z_R^{(1)} - z_R^{(2)}\right) \\
\widetilde{V}_{GR} V_{RR} \left(z_R^{(1)} - z_R^{(2)}\right) \\
\widetilde{V}_{BR} V_{RR} \left(z_R^{(1)} - z_R^{(2)}\right) 
\end{bmatrix}
\end{align}

Since $\widetilde{\mV}$ is a normalized version of $\mV$, it follows that $\widetilde{V}_{RR} = 1$. Therefore, we can find $\widetilde{V}_{GR}$ and $\widetilde{V}_{BR}$ as

\begin{equation}\label{eqn:vtilde}
    \widetilde{V}_{GR} = \frac{\Delta g_G}{\Delta g_R} \text{ , and }  \widetilde{V}_{BR} = \frac{\Delta g_B}{\Delta g_R}
\end{equation}

Similarly, the other elements of $\widetilde{\mV}$ can be found by using a pair of images for each of the green and blue color channels. To obtain the matrix $\widetilde{\mV}$, we need a total of 4 images (the first four images in \fref{fig:estimating_parameters_supp}). The gray scale image has a uniform input intensity of $85/255$ at all the pixels. The other three images are obtained by changing the input intensity for one color channel to $140/255$. 

\subsection*{Estimating $\calT^{-1}(\vg)$}
Using $\widetilde{\mV}^{-1}$, we can rewrite the image from
\begin{align*}
    \vg &= \underset{\widetilde{\mV}\mD}{\underbrace{\mV}}\vz + \vb
\end{align*}
to the modified equation:
\begin{align}
    \underset{\widetilde{\vg}}{\underbrace{\textcolor{blue}{\widetilde{\mV}^{-1}} \vg }}
    &=  \textcolor{blue}{\widetilde{\mV}^{-1}} \widetilde{\mV}\mD
    \underset{\calM(\vf)}{\underbrace{
    \vz}} + 
    \underset{\widetilde{\vb}}{\underbrace{\textcolor{blue}{\widetilde{\mV}^{-1}}\vb}}.
\end{align}

This new equation allows us to decouple the interaction between the color channels. Using the red channel as an example, we can write the red channel of $\widetilde{\vg}$ as
\begin{equation}
    \widetilde{g}_R = V_{RR} \calM(f_R)  + \widetilde{b}_R
\end{equation}

\begin{figure}[!]
\centering
\begin{tabular}{c}
\includegraphics[width=0.9\linewidth]{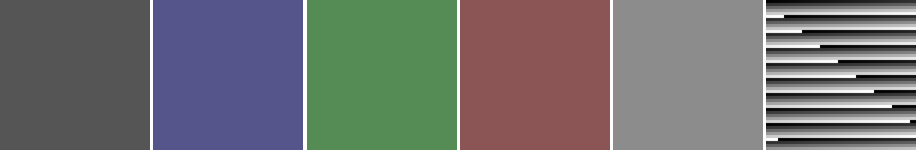}\\
(a) Input to the Projector \\
\includegraphics[width=0.9\linewidth]{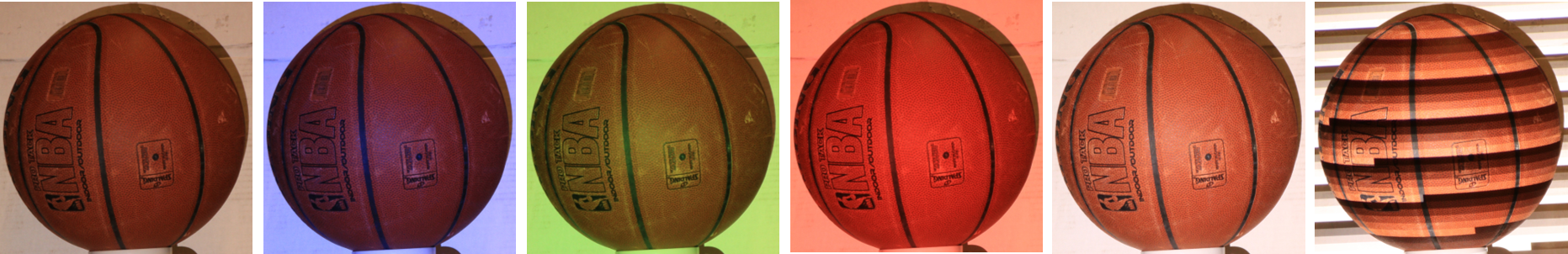}\\
(b) Captured images \\
\end{tabular}
\caption{Estimating the model parameters. For each scene we need to capture a total of 6 images to estimate the projector-camera forward model and inverse.}
\label{fig:estimating_parameters_supp}
\end{figure}

Our goal is to find the inverse transform $\calT^{-1}(\vg)$, i.e. we need to estimate the inputs $f_R$, $f_G$ and $f_B$ given $\vg$. Since we have already decoupled the interaction between the color channels, we have a single equation for each color channel. Therefore, from the above equation, we need to find $f_R$. 

One possible way to do this is to capture 256 images, each at different projector input intensity. This will give us a map between $\widetilde{g}_R$ and $f_R$ at each pixel for all possible values of $f_R$. Thus, for any given image, we can go to each pixel and estimate $f_R$, based on the $\widetilde{g}_R$. However, this process is time consuming. To speed up the process, we notice that the radiometric response 
$\calM_R(f_R)$ is fixed for a given projector and does not vary over the different pixel positions $x$. With that in mind, we define the following \textbf{invariant} which is a function of $f_R$

\begin{align}
\nonumber n_R \bydef & \frac{\widetilde{g}_R - \widetilde{g}^{(s)}_R }{\widetilde{g}^{(t)}_R - \widetilde{g}^{(s)}_R } \\
=& \frac{\calM_R(f_R) - \calM_R(f_R^{(s)})}{\calM_R(f_R^{(t)}) - \calM_R(f_R^{(s)})}
\end{align}
where $\widetilde{g}_R^{(s)}$ and $\widetilde{g}_R^{(t)}$ are from images obtained with uniform intensities of $s\cdot \mathbbm{1}$ and $t \cdot \mathbbm{1}$ respectively.

Notice that $n_R$ depends only on the red channel input $f_R$, and the radiometric function $\calM_R(\cdot)$ of the projector and does not depend on the scene at all. So, to learn the relationship, we do not need 256 different images, but just a single image which sweeps over all the 256 possible intensities. For this purpose, we use the last image in \fref{fig:estimating_parameters_supp}, which has the intensities varying from $0/255$ to $255/255$ repeatedly over the entire image. We can plot the corresponding $n_R$ vs $f_R$ at different pixels and fit a function $f_R = \rho_R(n_R)$. Similarly we can obtain the input intensities for the other channels too. So, we need a total of three images the one with varying intensities, ${\vg}^{(s)} = s\cdot \mathbbm{1}$, and ${\vg}^{(t)} = t \cdot \mathbbm{1}$. For ${\vg}^{(s)}$, we reuse the image with uniform intensity of $85/255$, and for ${\vg}^{(t)}$ we use an image with uniform intensity of $140/255$.

\begin{figure}[h]
    \centering
    \includegraphics[width=0.9\linewidth]{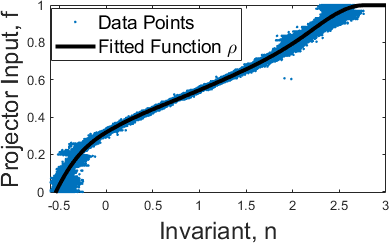}
    \caption{Fitting the function $\rho$. Given a image which needs to be captured by the camera. We find the invariant $\vn$. For each pixel, Based on the invariant $n$, we fit a function $\rho$ to estimate the projector input $f$. }
    \label{fig:rho}
\end{figure}

We fit the data using a monotonically increasing function to obtain $\vrho(\vn) = [\rho_R(n_R), \rho_G(n_G), \rho_B(n_B)]$. In \fref{fig:rho}, we show an example of the actual function $\rho_R(n_R)$ obtained from a real scene. 

Once we have the function $\vrho(\cdot)$, for any given scene pixel $\vg$, we first need to calculate the invariant $\vn$ and use the function $\vrho$ to estimate $\vf$. This can be written as,
\begin{equation} \label{eqn:inverse_Full}
    \mat{\widehat{f}} = 
    \underset{\calT^{-1}(\vg)}{\underbrace{
    \vrho\left(\left(\widetilde{\mV}^{-1}\vg - \vg^{(t)}\right)./\left(\vg^{(s)} - \vg^{(s)}\right) \right)}},
\end{equation}
where ($./$) denotes element-wise division. Algorithm~\ref{algo:inverse_model} summarizes the entire algorithm.

\begin{algorithm}
 1. Calculate $\widetilde{V}_{ij} \ \forall i,j \in \{R,G,B\},\ i\neq j$ according to \eref{eqn:vtilde}.\\
 2. Estimate function $\rho_R(\cdot)$, $\rho_G(\cdot)$, $\rho_B(\cdot)$ as shown in \fref{fig:rho}. \\
 3. For any given image $\vg$: Estimate $\mat{\widehat{f}}$ according to \eref{eqn:inverse_Full}.
\caption{Inverse $\calT^{-1}(\vg)$ \label{algo:inverse_model}}
\end{algorithm}

\subsection*{Estimating $\calT(\vf)$}
While we have discussed a way to find $\calT^{-1}(\vg)$, we have not explicitly found $\mV$ or $\calM(\cdot)$. So, we still cannot use the forward model $\calT(\vf)$. In this subsection we look at how we can find the forward model $\calT(\vf)$ without capturing any new images. We use only the six images we used for finding $\calT^{-1}(\vg)$ to estimate the forward model. 

First, notice that the invariant 
\begin{equation}
n_R = \frac{\calM_R(f_R) - c_R }{d_R},    
\end{equation}
where $c_R = \calM_R(f_R^{(s)})$ and $d_R = \calM_R(f_R^{(t)}) - \calM_R(f_R^{(s)})$ are constants if $f_R^{(s)}$ and $f_R^{(t)}$ are fixed. So, the radiometric function can be written as 
\begin{equation}
    \calM_R(f_R) = n_R\ d_R + c_R = \rho_R^{-1}(f_R)\ d_R + c_R
\end{equation}
Thus, \eref{eqn:Deltas} can be re-written as
\begin{align}\label{eqn:12}
\begin{bmatrix}
g_R^{(1)} - g_R^{(2)}\\
g_G^{(1)} - g_G^{(2)}\\
g_B^{(1)} - g_B^{(2)}
\end{bmatrix} = 
\begin{bmatrix}
V_{RR}\  d_R \  \left(\rho_R^{-1}(f_R^{(1)}) - \rho_R^{-1}(f_R^{(2)})\right)\\
V_{GR}\  d_R \  \left(\rho_R^{-1}(f_R^{(1)}) - \rho_R^{-1}(f_R^{(2)})\right)\\
V_{BR}\  d_R \  \left(\rho_R^{-1}(f_R^{(1)}) - \rho_R^{-1}(f_R^{(2)})\right)
\end{bmatrix}
\end{align}
Rearranging the terms we get 
\begin{align}\label{eqn:MRRbR}
\begin{bmatrix}
V_{RR}\  d_R\\
V_{GR}\  d_R\\
V_{BR}\  d_R
\end{bmatrix} = 
\begin{bmatrix}
\Delta g_R / \left\{\left(\rho_R^{-1}(f_R^{(1)}) - \rho_R^{-1}(f_R^{(2)})\right)\right\}\\
\Delta g_G /  \left\{\left(\rho_R^{-1}(f_R^{(1)}) - \rho_R^{-1}(f_R^{(2)})\right)\right\}\\
\Delta g_R / \left\{\left(\rho_R^{-1}(f_R^{(1)}) - \rho_R^{-1}(f_R^{(2)})\right)\right\} 
\end{bmatrix}
\end{align}
We can obtain the other 6 terms similarly. Now, let us look at the expression for $g_R$.
\begin{eqnarray}
\nonumber g_R =& V_{RR}\ d_R\ \rho_R^{-1}(f_R) + V_{RG}\ d_G\ \rho_G^{-1}(f_G) +\\
\nonumber & V_{RB}\ d_B\ \rho_R^{-1}(f_B) + V_{RR}\ c_R + V_{RG}\ c_G + \\ &V_{RB}\ c_B + b_R 
\end{eqnarray}
\\
Notice that $V_{RR}\ d_R\ \rho_R^{-1}(f_R) + V_{RG}\ d_G\ \rho_G^{-1}(f_G) + V_{RB}.d_B.\rho_R^{-1}(f_B) $ can be obtained using \eref{eqn:MRRbR}. The rest of the terms are constant for a given pixel. We denote them by $$\bar{b}_R =  V_{RR}\ c_R + V_{RG}\ c_G + V_{RB}\ c_B + b_R .$$

We can obtain $\bar{b}_R$ from any image as  
\begin{eqnarray}\label{eqn:br}
\nonumber \bar{b}_R = g_R - (V_{RR}\ d_R\ \rho_R^{-1}(f_R) + V_{RG}\ d_G\ \rho_G^{-1}(f_G) \\
+ V_{RB}\ d_B\ \rho_R^{-1}(f_B))
\end{eqnarray}

Similarly we can obtain $\bar{b}_G$ and $\bar{b}_B$. Then, for any given input pixel $\vf$, the corresponding pixel in the captured image $\vg$ can be estimated as

\begin{align}\label{eqn:vg}
\begin{bmatrix}
\widehat{g}_{R}\\
\widehat{g}_{G}\\
\widehat{g}_{B} 
\end{bmatrix} &=
\begin{bmatrix}
V_{RR}\ d_R & V_{RG}\ d_G & V_{RB}\ d_B \\
V_{GR}\ d_R & V_{GG}\ d_G & V_{GB}\ d_B\\
V_{BR}\ d_R & V_{BG}\ d_G & V_{BB}\ d_B 
\end{bmatrix} 
\begin{bmatrix}
\rho_R^{-1}(f_R)\\
\rho_G^{-1}(f_G)\\
\rho_B^{-1}(f_B) 
\end{bmatrix} \notag \\
&\qquad +
\begin{bmatrix}
\bar{b}_R\\
\bar{b}_G\\
\bar{b}_B
\end{bmatrix},
\end{align}
or simply
\begin{equation}
    \widehat{\vg} = \calT(\vf).
\end{equation}
The first matrix can be obtained using \eref{eqn:MRRbR}. We know what $\vrho^{-1}(\cdot)$ is. $\bar{\vb}$ can be obtained using \eref{eqn:br}. Thus, for any given input $\vf$, we can estimate the image $\vg$. Algorithm~\ref{algo:forward_model} summarizes the entire algorithm.

\begin{algorithm}
 1. Calculate $V_{ij} d_j,\ \forall i,j \in \{R,G,B\}$ according to \eref{eqn:MRRbR}.\\
 2. Calculate $b_R,\ b_G,\ b_B$ according to \eref{eqn:br}.\\
 3. Estimate the functions $\rho_R^{-1}(\cdot),\ \rho_G^{-1}(\cdot),\ \rho_B^{-1}(\cdot)$, the inverse of the function obtained in \fref{fig:rho}. \\
 4. For any given input $\vf$: Estimate $\hat{\vg}$ according to \eref{eqn:vg}.
\caption{Forward model $\calT(\vf)$ \label{algo:forward_model}}
\end{algorithm}

\begin{figure}[h]
    \centering
    \includegraphics[width=0.9\linewidth]{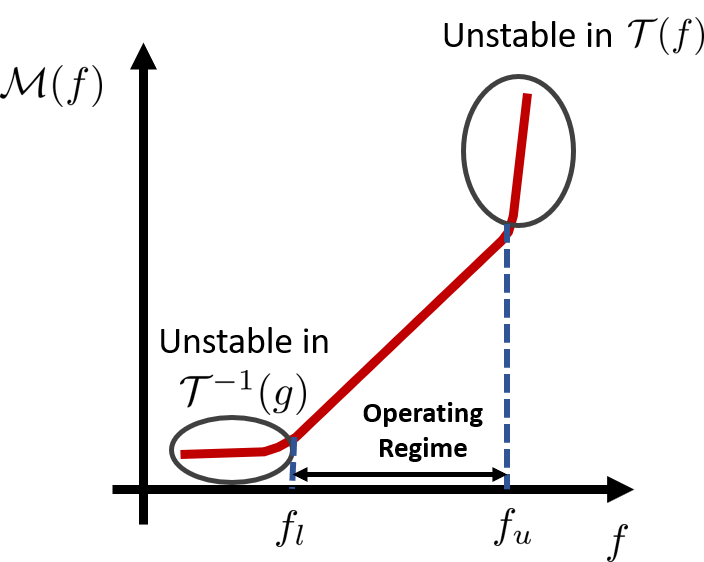}
    \vspace{-2ex}
    \caption{Operating regime of OPAD. }
    \label{fig:operating_reg}
\end{figure}

\begin{figure*}[t]
\centering
\begin{tabular}{cccccccc}
True Obj. & \hspace{-2.0ex} Tgt. apprnce  & \hspace{-2.0ex} Illumination & \hspace{-2.0ex} Captured &\hspace{-2.0ex}True Obj. & \hspace{-2.0ex} Tgt. apprnce  & \hspace{-2.0ex} Illumination & \hspace{-2.0ex} Captured \\
\includegraphics[width=0.12\linewidth]{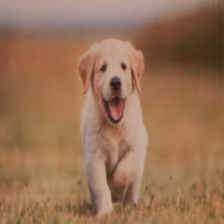}&
\hspace{-2.0ex}\includegraphics[width=0.12\linewidth]{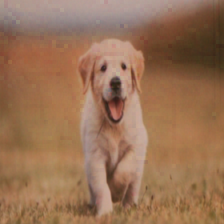}&
\hspace{-2.0ex}\includegraphics[width=0.12\linewidth]{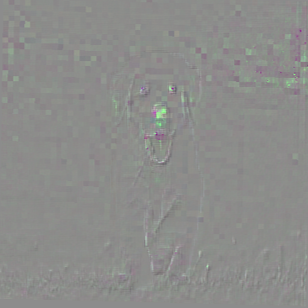}&
\hspace{-2.0ex}\includegraphics[width=0.12\linewidth]{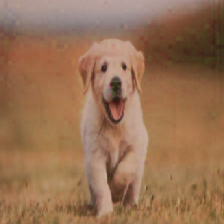} &
\hspace{-2.0ex}\includegraphics[width=0.12\linewidth]{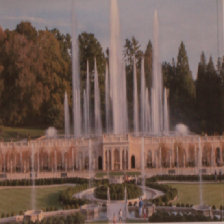}&
\hspace{-2.0ex}\includegraphics[width=0.12\linewidth]{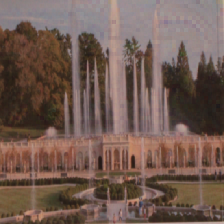}&
\hspace{-2.0ex}\includegraphics[width=0.12\linewidth]{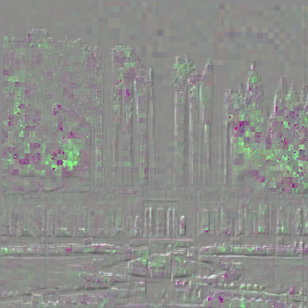}&
\hspace{-2.0ex}\includegraphics[width=0.12\linewidth]{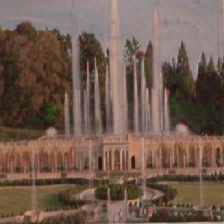}\\
Gold. Retr. & \hspace{-2.0ex}Rain Barrel &\hspace{-2.0ex} &\hspace{-2.0ex} Rain Barrel & \hspace{-2.0ex}Fountain & \hspace{-2.0ex}Lakeside &\hspace{-2.0ex} & \hspace{-2.0ex}Lakeside \\
\includegraphics[width=0.12\linewidth]{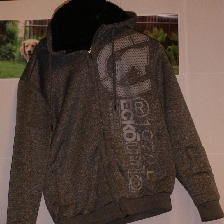}&
\hspace{-2.0ex}\includegraphics[width=0.12\linewidth]{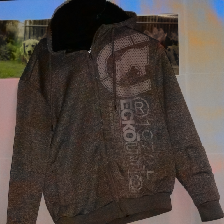}&
\hspace{-2.0ex}\includegraphics[width=0.12\linewidth]{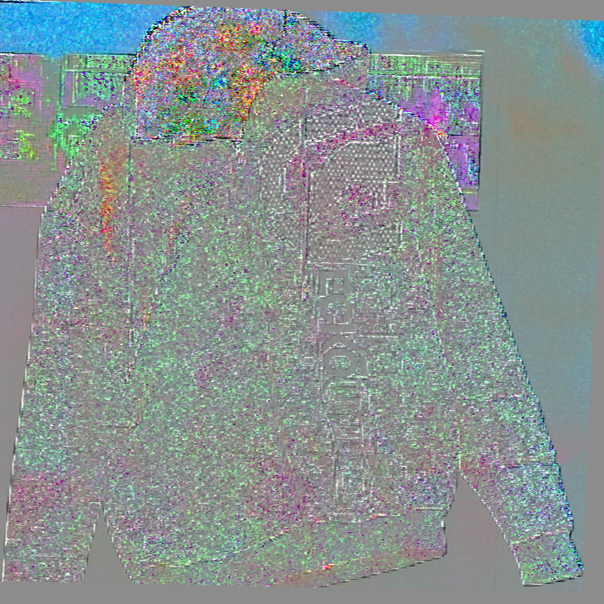}&
\hspace{-2.0ex}\includegraphics[width=0.12\linewidth]{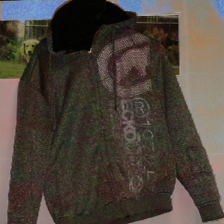}&
\hspace{-2.0ex} \includegraphics[width=0.12\linewidth]{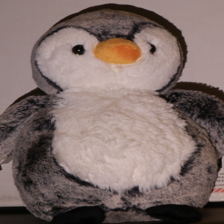}&
\hspace{-2.0ex}\includegraphics[width=0.12\linewidth]{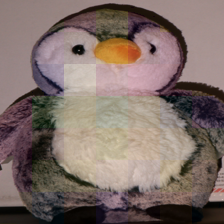}&
\hspace{-2.0ex}\includegraphics[width=0.12\linewidth]{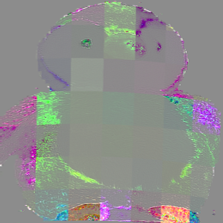}&
\hspace{-2.0ex}\includegraphics[width=0.12\linewidth]{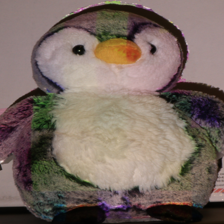}\\
Cardigan & \hspace{-2.0ex}Poncho &\hspace{-2.0ex} &\hspace{-2.0ex} Poncho &\hspace{-2.0ex} Teddy &\hspace{-2.0ex} Wool &\hspace{-2.0ex} &\hspace{-2.0ex} Wool
\end{tabular}
\caption{Additional Experiments. We use a white-box on `golden retriever' and `cardigan', and black-box attack on the `fountain' and `teddy'. `Cardigan' is attacked using colorization attack \cite{bhattad_19_color_att}. Other three images use PGD \cite{madry2017towards} attacks. }
\label{fig:additional Experiments}
\end{figure*}

\begin{table*}[ht]
    \centering
    \begin{tabular}{cccccccc}
        \hline\hline
         Image & Algorithm& Type & Generated on & Tested on & Block Size & St.-size($\alpha$)& Norm. $\ell_2 $ Dist.\\
         \hline
         \multicolumn{8}{c}{\textbf{Supplementary Document}}\\
         Golden Retriever& PGD ($\ell_2$) & White-box & VGG-16 & VGG-16 & $4\times4$ &0.5 &3.1/255\\
         Fountain & PGD ($\ell_2$) & Black-box & VGG-16 & Resnet-50 & $4\times4$ &0.5&4.4/255\\
         Cardigan  & Colorization & White-box & VGG-16 & VGG-16 &  $1\times1$ &N/A& 6.9/255\\
         Teddy & PGD ($\ell_2$) & Black-box & Resnet-50 & VGG-16 &$32\times32$ &0.5 & 3.9/255 \\
         \hline
         \multicolumn{8}{c}{\textbf{Main Paper}}\\
          Cardigan (Fig. 8)  & PGD  ($\ell_\infty$) & White-box          & VGG-16 & VGG-16 & $8\times8$ &0.05 & 5.8/255\\
         Basketball(Fig. 8)   & PGD ($\ell_\infty$) & White-box & Resnet-50 & Resnet-50 &$8\times8$ &0.05&3.2/255 \\
         Coffee Mug(Fig. 8)   & PGD ($\ell_2$) & White-box  & VGG-16 & VGG-16 &$8\times8$ &0.5&4.1/255 \\
         Teddy(Fig. 8)        & PGD ($\ell_2$) & White-box  & Resnet-50 & Resnet-50 &$8\times8$ &0.5 & 4.3/255 \\
         Book (Fig. 9)  & PGD ($\ell_2$) & White-box & VGG-16 & VGG-16 & $4\times4$ & 1 & 3.6/255\\
         Stop Sign (Fig. 12) & PGD ($\ell_\infty$) & White-box & GTSRB-CNN & GTSRB-CNN &$1\times1$  & 0.05 & 2.1/255\\
         Stop Sign (Fig. 13) & PGD ($\ell_\infty$) & White-box & GTSRB-CNN & GTSRB-CNN &$1\times1$  & 0.05 & 3.2/255 \\ 
         \hline
    \end{tabular}
    \caption{Details about the experiments conducted. }
    \label{tab:experiments}
\end{table*}

\subsection*{Operating regime of OPAD}
In \fref{fig:operating_reg}, we show how it is not possible to use all possible input intensities for the projector input. The forward model becomes unstable at higher intensities because even small change in input $f$ introduces large change in the in the radiometric response $\calM(f)$. Similarly the inverse function becomes unstable in the darker region. So, we choose to operate OPAD in the middle where the gradient is not small or not too large. Notice that the the function $\rho^{-1}(\cdot)$ is linearly related to $\calM(\cdot)$. So, we can find the middle region from the function $\rho^{-1}(\cdot)$. Changing the range of possible projector input intensities from $[0,1]$ to some other $[f_l,f_u]$ does not affect the OPAD attack algorithm defined in the main paper. The only change will be - instead of clipping the input intensities to [0,1], we now have to clip them in $[f_l,f_u]$.

\section*{Additional details experiments}
We demonstrated only white-box projected gradient descent (PGD) \cite{madry2017towards} and fast gradient sign method (FGSM) \cite{goodfellow_14_explaining_adversarial} attacks in the main document. In \fref{fig:additional Experiments}, we perform black-box, and white-box attacks using PGD and colorization methods \cite{bhattad_19_color_att} using different block sizes. Details about these experiments and the attacks performed in the main paper can be found in \tref{tab:experiments}.  More detailed description about the experiments performed can be found in this section below.

\begin{itemize}
    \item \textbf{Attack algorithm.} The proposed OPAD system is tested on the following types of attack algorithms: fast gradient sign method (FGSM) \cite{goodfellow_14_explaining_adversarial}, projected gradient descent (PGD) \cite{madry2017towards}, and colorization \cite{bhattad_19_color_att}. We use two different versions of PGD, one with $\ell_\infty$ constraints and one with $\ell_2$ constraints. The PGD algorithms are run for 20 iterations in all experiments.
    
    \item \textbf{Attack type.} All the attacks reported in the paper are targeted attacks. Un-targeted attacks are significantly easier to achieve because one just needs to increase the perturbation magnitude. As such, we decide to skip un-targeted attacks in this paper. For targeted attacks, we demonstrate both white-box and black-box attacks. In the white-box attacks, we assume that we have access to the model parameters of the classifiers. Therefore, we can use the same classifier when generating the attack. In the black-box attacks, we assume that we do not have access to the classifiers and so we need to use a different classifier to generate the attack.
    
    \item \textbf{Types of networks.} For the traffic sign dataset, we use the neural network trained on the  German Traffic Sign Recognition Benchmark (GTSRB) dataset \cite{stallkamp2012man}, as trained in \cite{eykholt_18_stopsign_phy}. We refer to this network as GTSRB-CNN. For the ImageNet datasets, we use two different kinds of networks - VGG16 \cite{simonyan2014very}, and Resnet-50 \cite{he2016deep}. Both VGG-16 and ResNet-50 are the pre-trained models provided by the original authors.
    
    \item \textbf{Block size.} The optics of the projector and camera may cause blur to some of the attack patterns. In those cases, we attack a block of pixels instead of individual pixels, where multiple pixels share the same update. For example, we can use a block size of $4 \times 4$ with PGD ($\ell_2$) attack. For each $4\times4$ block of pixels, we calculate the average of gradients, and use this average to update all the pixels in the block. In the experiments, we have used block sizes of $1\times1$, $4\times4$, $8\times8$, and $32\times32$.
    
    \item \textbf{Step size.} For PGD and FGSM, the step-size for each iteration of the attack algorithm depends on the constraint imposed. The different step-size used for different images can be found in \tref{tab:experiments}. 
    
    \item \textbf{Details about Section 5.2} In section 5.2, we attacked the object `book' by targeting 15 random classes from the ImageNet dataset \cite{imagenet_cvpr09}. The fifteen classes are `American alligator', `black swan', `sea lion', `Tibetan terrier', `Siberian husky', `Tiger beetle', `Capra ibex', `Academic gown', `Bobsleigh', `Cliff dwelling', `Espresso maker', `Claw', `Microphone', `Comic-book', and `Pretzel'. \cite{CVPRW_attack} was developed for face recognition. It attacks a real face by minimizing the cosine distance between the face embedding of a given person's image and the captured image. We replace this with the loss function of the image classifier. Thus, the only difference between the proposed method and the modified \cite{CVPRW_attack} is the camera-projector calibration. \cite{CVPRW_attack} models the camera-projector model as a global color correction, without taking into account the spatially varying spectral response of the scene. OPAD takes this into consideration.  
    
\end{itemize}

{
\bibliographystyle{ieee_fullname}
\bibliography{egbib}
}

\end{document}

%% file: commands.tex
\usepackage{multirow}
\usepackage{times}
\usepackage{amsmath,amssymb}
\usepackage{algorithm}
\usepackage{epstopdf}
\usepackage{subfigure}
\usepackage[dvipsnames]{xcolor}
\usepackage{algorithm}
\usepackage{algorithmic}
\usepackage{cite}

\usepackage{bbm}

{\begin{list}               
    {$\bullet$ \hfill}{
        \setlength{\leftmargin}{\parindent}
        \setlength{\parsep}{0.04\baselineskip}
        \setlength{\itemsep}{0.5\parsep}
        \setlength{\labelwidth}{\leftmargin}
        \setlength{\labelsep}{0em}}
    }
{\end{list}}

\providecommand{\eref}[1]{\eqref{#1}}  
\providecommand{\cref}[1]{Chapter~\ref{#1}}

\providecommand{\fref}[1]{Figure~\ref{#1}}
\providecommand{\tref}[1]{Table~\ref{#1}}

\providecommand{\R}{\ensuremath{\mathbb{R}}}

\providecommand{\bydef}{\overset{\text{def}}{=}}

\renewcommand{\vec}[1]{\ensuremath{\boldsymbol{#1}}}
\providecommand{\mat}[1]{\ensuremath{\boldsymbol{#1}}}

\providecommand{\calL}{\mathcal{L}}
\providecommand{\calM}{\mathcal{M}}

\providecommand{\calT}{\mathcal{T}}

\providecommand{\mD}{\mat{D}}

\providecommand{\mV}{\mat{V}}

\providecommand{\vb}{\vec{b}}
\providecommand{\vc}{\vec{c}}

\providecommand{\vf}{\vec{f}}
\providecommand{\vg}{\vec{g}}

\providecommand{\vn}{\vec{n}}

\providecommand{\vz}{\vec{z}}

\providecommand{\vdelta}{\vec{\delta}}

\providecommand{\veta}{\vec{\eta}}
\providecommand{\vtheta}{\vec{\theta}}

\providecommand{\vrho}{\vec{\rho}}

\newcommand{\argmax}[1]{\mathop{\underset{#1}{\mbox{argmax}}}}
